\documentclass[preprint,12pt]{elsarticle}

\usepackage{color,soul}

\usepackage{amssymb}
\usepackage{amsmath}

\usepackage{tabularx} 
\usepackage{subfigure}
\usepackage{multicol}
\usepackage{multirow}
\usepackage{url}
\usepackage{booktabs}
\usepackage{hyperref}
\usepackage{pifont} 
\usepackage{array}
\usepackage{graphicx}
\usepackage{subcaption}
\usepackage{booktabs}
\usepackage{multirow}
\usepackage{graphicx}
\usepackage{array}
\usepackage{siunitx}  
\usepackage{xcolor}
\usepackage{makecell}
\usepackage{capt-of}

\journal{Nuclear Physics B}

\begin{document}

\begin{frontmatter}



\title{\textsc{EPPCMinerBen}: A Novel Benchmark for Evaluating Large Language Models on Electronic Patient-Provider Communication via the Patient Portal}




\author[label1]{Samah Jamal Fodeh\corref{cor1}}
\ead{samah.fodeh@yale.edu}

\author[label1]{Yan Wang}
\author[label1]{Linhai Ma}
\author[label1]{Srivani Talakokkul}
\author[label2]{Jordan M. Alpert}
\author[label3]{Sarah Schellhorn}

\cortext[cor1]{Corresponding author}

\affiliation[label1]{organization={Department of Emergency Medicine, Yale School of Medicine},
            addressline={464 Congress Ave}, 
            city={New Haven},
            postcode={06519}, 
            state={CT},
            country={USA}}

\affiliation[label2]{organization={Cleveland Clinic Lerner College of Medicine of Case Western Reserve University, Cleveland Clinic},
            addressline={9501 Euclid Ave}, 
            city={Cleveland},
            postcode={44195}, 
            state={OH},
            country={USA}}

\affiliation[label3]{organization={Medical Oncology, Yale School of Medicine},
            addressline={300 George St}, 
            city={New Haven},
            postcode={06510}, 
            state={CT},
            country={USA}}

\begin{abstract}
Effective communication in health care is critical for treatment outcomes and adherence. With patient-provider exchanges shifting to secure messaging, analyzing electronic patient-communication (EPPC) data is both essential and challenging. We introduce \textsc{EPPCMinerBen}, a benchmark for evaluating LLMs in detecting communication patterns and extracting insights from electronic patient-provider messages.
\textsc{EPPCMinerBen} includes three sub-tasks: Code Classification, Subcode Classification, and Evidence Extraction. Using 1,933 expert-annotated sentences from 752 secure messages of the patient portal at Yale New Haven Hospital, it evaluates LLMs on identifying communicative intent and supporting text. Benchmarks span various LLMs under zero-shot and few-shot settings, with data to be released via the NCI Cancer Data Service.
Model performance varied across tasks and settings. Llama-3.1-70B led in evidence extraction (F1: 82.84\%) and performed well in classification. Llama-3.3-70b-Instruct outperformed all models in code classification (F1: 67.03\%). DeepSeek-R1-Distill-Qwen-32B excelled in subcode classification (F1: 48.25\%), while sdoh-llama-3-70B showed consistent performance. Smaller models underperformed, especially in subcode classification ($>$30\% F1). Few-shot prompting improved most tasks.
Our results show that large, instruction-tuned models generally perform better in \textsc{EPPCMinerBen} tasks, particularly evidence extraction while smaller models struggle with fine-grained reasoning. \textsc{EPPCMinerBen} provides a benchmark for discourse-level understanding, supporting future work on model generalization and patient–provider communication analysis.
\end{abstract}




\end{frontmatter}



\section{Introduction}
Effective management of chronic conditions like cancer requires not only access to medication but also consistent adherence, particularly for oral chemo-therapies. In the U.S., nearly two million people are diagnosed with cancer annually~\cite{siegel2022cancer}, and adherence to oral chemotherapy can significantly improve outcomes and reduce healthcare costs. However, adherence remains suboptimal, with rates falling to 20\% in some populations~\cite{thomas2019challenges}, often due to challenges in patient-provider communication (PPC), such as misunderstandings about dosing, side effects, and treatment schedules. As many activities related to cancer care are now handled using electronic patient-provider communication (EPPC) via secure patient portals~\cite{neeman2021attitudes}, these platforms have become critical for supporting adherence and overall care. EPPC systems generate large volumes of unstructured secure messages (SM) that capture patients’ questions, concerns, and treatment updates, offering valuable opportunities to identify communication breakdowns, uncover behavioral barriers to adherence, and enable personalized interventions at scale.

 Despite the potential of EPPC data, there remains a research gap in leveraging it to meaningfully characterize EPPC. First, most studies rely on manual coding or rule-based techniques~\cite{shimada2013patient,zhou2007patient,price2018dose},  which are costly and lack scalability and robustness ~\cite{ma2022regularization,ma2023improving,ma2023increasing}. Second, while some studies apply machine learning~\cite{sulieman2017classifying,cronin2017comparison,wang2018bidirectional}, they often suffer from methodological constraints, such as the inability to capture multi-label or bidirectional communications focusing on patient-authored SM, overlooking provider responses and the interactive context essential to understanding EPPC. Third, although recent natural language processing (NLP) resources ~\cite{zhang2022cblue,liu2022meddg,saley2024meditod,yan2022remedi,ahsan2021mimic,lelkes2023sdoh} have advanced biomedical dialogue and Social Determinants of Health (SDoH) extraction, they are not tailored to secure messaging platforms. Finally, progress in characterizing EPPC is further hindered by limited access to de-identified, richly annotated EPPC data and the lack of benchmarks that reflect its relational and contextual complexity. These gaps underscore the need for advanced language models and purpose-built benchmarks to support scalable, nuanced EPPC analysis.

To address these challenges, our objective is to develop  \textsc{EPPCMinerBen}we developed EPPCMinerBen, a benchmark designed to extract communication patterns from secure messages (SM) and capture the interactive dynamics between patients and providers, rather than focusing solely on patient-initiated engagement. The benchmark highlights a diverse range of EPPC codes—including clinical information seeking and sharing, socio-emotional expressions, partnership-building behaviors, and shared decision-making. We constructed an annotated dataset of patient–provider SM and manually labeled each message with the corresponding EPPC codes. This dataset enabled  \textsc{EPPCMinerBen}  to systematically evaluate  large language models in analyzing patient-provider interactions with both contextual clarity and relational nuance. This benchmark ultimately aims to support researchers and NLP practitioners in developing tools that enhance bidirectional communication, promote more personalized and equitable healthcare, and inform future clinical NLP applications such as communication training tools.

EPPCMinerBen offers interdisciplinary value by bridging computational linguistics, clinical communication research, and healthcare practice. By providing a structured framework for analyzing patient–provider communication, the benchmark supports methodological advances in NLP while also enabling clinicians, communication scientists, and informatics researchers to study communication quality, clinical intent, and patient needs at scale. This integrated perspective broadens the benchmark’s relevance beyond NLP and positions it as a resource with meaningful clinical implications for both computational and clinical communities. Systematically characterizing EPPC can help identify unmet informational, emotional, or logistical needs that may otherwise go unrecognized in routine care. These insights can support clinicians in tailoring their communication strategies to individual patients and in prioritizing messages that require timely follow-up, thereby improving adherence, care coordination, and patient satisfaction.

\begin{figure}
    \centering
    \includegraphics[width=1\linewidth]{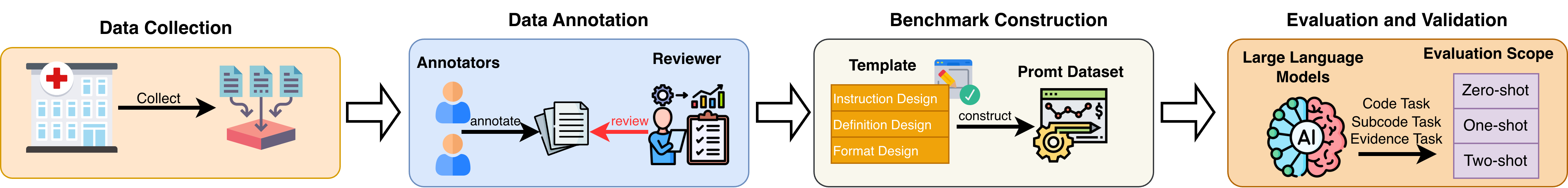}
    \caption{{Overall Pipeline of EPPCMinerBen}}
    \label{fig:flowchart}
\end{figure}

All in all, our contributions are as follows:
(1) This work introduces \textsc{EPPCMinerBen}, a benchmark specifically constructed to analyze EPPC within large-scale secure messaging data, addressing a critical gap left by existing clinical dialogue datasets.
(2) This work is based on a richly annotated, de-identified EPPC dataset that captures diverse relational communication patterns, including  information seeking/sharing, socio-emotional expressions, partnership-building, and shared decision-making. The dataset provides a strong foundation for analyzing the interactive dynamics of patient-provider communication and supports the development of models that reflect real-world conversational complexity. 
(3) This work further proposes a set of tailored instruction prompts designed to guide large language models in interpreting and reasoning over EPPC data. Figure~\ref{fig:flowchart} illustrates the pipeline of EPPCMinerBen benchmark. 

The remainder of the paper is organized as follows. We begin with reporting the related work in the literature followed be the materials and methods in which we present the set of tasks designed to evaluate large language models followed by a description of our experimental setup and evaluation methodology. Next we introduce the EPPCMinerBen dataset, describing its construction and the annotation of patient–provider communication. Then we present the results, and analyze model performance, and Finally discuss the broader implications of our findings for computational linguistics, clinical communication research, and healthcare practice followed by our conclusion.

\section{Related Work}\label{sec2}

\begin{table*}[t!]
\centering
\tiny
\setlength{\tabcolsep}{2pt}
\renewcommand{\arraystretch}{0.85}

\caption{Comparison between EPPCMinerBen and existing biomedical NLP benchmarks. A check mark (\(\checkmark\)) indicates strong support, a cross mark (\ding{55}) indicates no support, and a triangle $\triangle$ indicates partial or limited support. \label{tab:ben_compare}}

\begin{tabular}{
>{\centering\arraybackslash}m{1.8cm}
>{\centering\arraybackslash}m{1.6cm}
>{\centering\arraybackslash}m{2.0cm}
>{\centering\arraybackslash}m{1.9cm}
>{\centering\arraybackslash}m{1.3cm}
>{\centering\arraybackslash}m{1.3cm}
>{\centering\arraybackslash}m{1.3cm}
>{\centering\arraybackslash}m{1.3cm}
}
\hline
\textbf{Benchmark} &
\textbf{Domain} &
\textbf{Focus} &
\textbf{Data Source} &
\textbf{Relational/ Socio-Emotional} &
\textbf{Bidirecti\-onal Interaction} &
\textbf{Supports Multi-Label Coding} &
\textbf{Tailored for Secure Messaging} \\
\hline

CBLUE &
Chinese biomedical &
NER, RE, classification &
Clinical text\& dialogues &
\ding{55} & \ding{55} & \ding{55} & \ding{55} \\
\hline

MedDG &
Chinese medical dialogue &
Diagnosis, symptom inquiry &
Simulated/ semi-structured dialogues &
\ding{55} & \ding{55} & \ding{55} & \ding{55} \\
\hline

ReMeDi &
English medical dialogue &
Medication intent, treatment reasoning &
Clinical conversations (non-SM) &
$\triangle$ & $\triangle$ & \ding{55} & \ding{55} \\
\hline

MediTOD &
English history-taking &
Symptom elicitation, clinical reasoning &
Simulated structured dialogues &
\ding{55} & \ding{55} & \ding{55} & \ding{55} \\
\hline

BLURB &
Biomedical NLP &
NER, RE, QA, summarization &
Biomedical literature &
\ding{55} & \ding{55} & \ding{55} & \ding{55} \\
\hline

\textbf{EPPCMiner\-Ben (ours)} &
U.S. secure messaging (EPPC) &
Communication functions, relational behaviors, adherence cues &
De-identified SM from patient portals &
\checkmark & \checkmark & \checkmark & \checkmark \\
\hline

\end{tabular}

\label{tab:narrow_header}
\end{table*}

 NLP has enabled scalable analysis of clinical texts across tasks such as named entity recognition, relation extraction, document classification, and question answering. Several benchmarks have been developed to evaluate NLP systems in biomedical and clinical domains; however, few are tailored to the relational and interactive nature of EPPC.

Existing benchmarks like CBLUE~\cite{zhang2022cblue}, MedDG~\cite{liu2022meddg}, and ReMeDi~\cite{yan2022remedi} focus on Chinese biomedical or task-oriented dialogues, while MediTOD~\cite{saley2024meditod} provides structured English history-taking data. Similarly, BLURB~\cite{gu2021domain} and recent large-scale evaluations~\cite{chen2025benchmarking} target traditional biomedical tasks. However, these resources overlook the dynamic, relational, and patient-centered nature of secure messaging, which features casual, emotional, and context-dependent communication distinct from clinical documentation or goal-oriented interactions. Table~\ref{tab:ben_compare} highlights these gaps by comparing \textsc{EPPCMinerBen} with existing biomedical NLP benchmarks. Prior datasets focus on structured clinical dialogues, task-oriented interactions, or biomedical literature and therefore lack relational or socio-emotional content, do not support multi-label communication coding, and are not tailored to secure messaging. None of these benchmarks model the bidirectional, informal, and context-dependent nature of real-world EPPC. \textsc{EPPCMinerBen} directly addresses these limitations by providing the first benchmark designed to capture the interactive and relational dynamics of patient-provider messaging.

Domain-specific models have significantly advanced biomedical NLP by adapting language representations to clinical and scientific texts. Early efforts like BioWordVec~\cite{zhang2019biowordvec} and BioSentVec~\cite{chen2019biosentvec} improved medical term handling through specialized embeddings. Transformer-based models such as BioBERT~\cite{lee2020biobert}, PubMedBERT~\cite{gu2021domain}, and SciFive~\cite{phan2021scifive} achieved state-of-the-art results in biomedical tasks via pretraining on domain-specific corpora. Generative models like BioGPT~\cite{luo2022biogpt} and BioBART~\cite{yuan2022biobart} further expanded capabilities to generation and summarization. However, these models have not been systematically evaluated on the interactive, informal, and emotionally nuanced nature of patient-provider communication in secure messaging contexts.

Recent advances in large language models (LLMs) have significantly enhanced natural language understanding and generation, largely driven by instruction tuning and model alignment~\cite{ouyang2022training,wang2025large,jiang2024mixtral}. The emergence of open-access models like LLaMA~\cite{touvron2023llama}, alongside proprietary frontier models such as GPT~\cite{achiam2023gpt}, has greatly expanded the LLM ecosystem and spurred a wide range of applications across healthcare, education, and scientific research. However, current LLMs have not been systematically evaluated for their ability to understand the nuanced relational dynamics of EPPC, including emotional tone, identifying social or contextual barriers, and tracking shared decision-making over time. 

As emphasized by Lee et al.~\cite{lee2023ai}, LLMs hold significant promise in improving patient engagement, adherence, and personalized care. \textsc{EPPCMinerBen} builds on this vision by introducing the first benchmark explicitly designed to evaluate how LLMs interpret and reason about asynchronous real-world patient-provider messaging, capturing the complexities of social, clinical and emotional exchange in a secure communication environment.

\section{Materials and Methods}\label{sec3}

\subsection{Task Formulation}\label{subsec2}

\begin{figure}[h]
\centering
\includegraphics[width=0.8\textwidth]{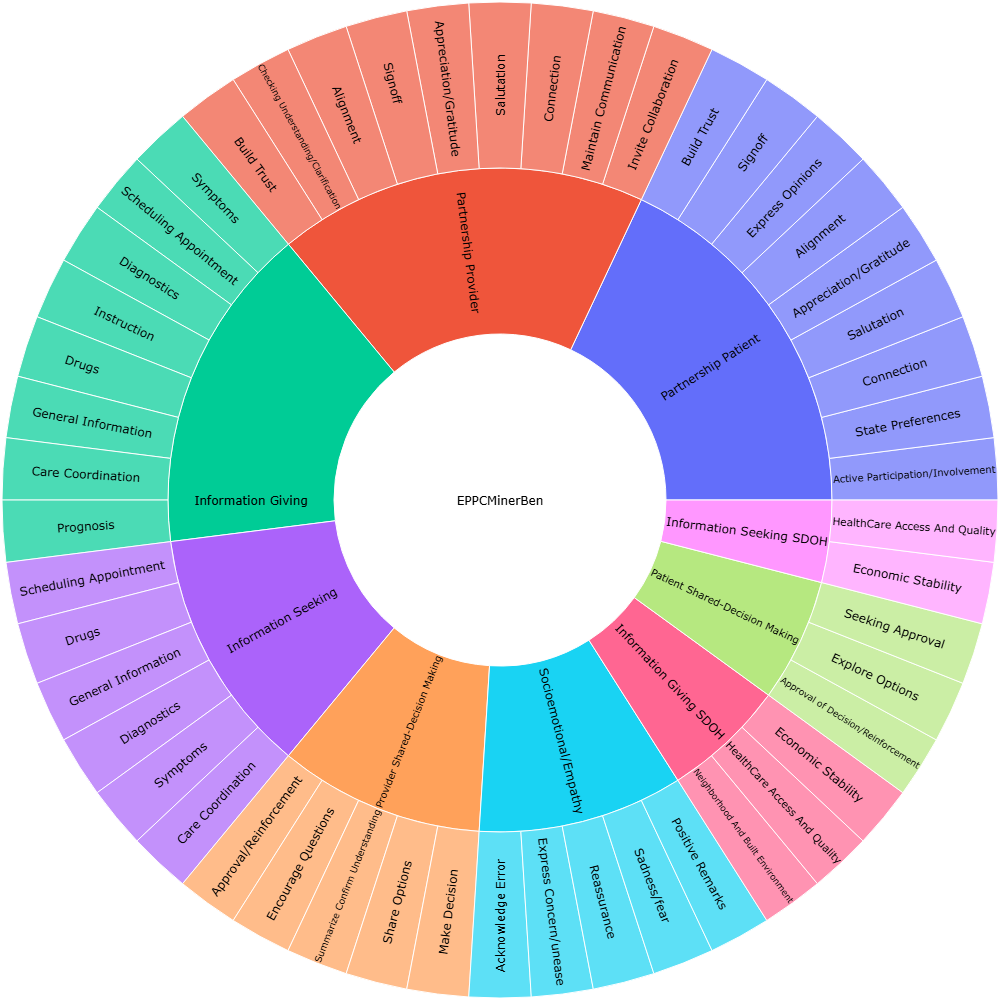}
\caption{Code and Subcode categories.}\label{fig_1}
\end{figure}

In this study, we formulate {\textsc{EPPCMinerBen}} as an information extraction task that involves identifying and extracting evidence spans from secure patient-provider messages to capture the interactive dynamics between the two parties. To represent these dynamics more comprehensively, we organize them by semantic granularity, as illustrated in Figure~\ref{fig_1}: coarse-grained categories (in 2nd layer) are defined as \textit{Codes}, and fine-grained categories (in 3rd layer) as \textit{Subcodes}. 
%
Specifically, assuming a message $M=\{s_1,s_2,\cdots,s_n\}$ containing $n$ sentences, given a current sentence $s_i$ in the message $M$ along with the context information $C^{(i)}=\{s_{i-1},s_{i+1}\}$, where $s_{i-1}$ is the previous sentence of $s_i$ and $s_{i+1}$ is the next sentence of $s_i$. The goal of this benchmark study is to leverage LLMs to identify the set of triplets $\{code, subcode, evidence\}$ for the current sentence $s_i$ through the three sub-tasks: Code Classification, Subcode Classification, and Evidence Extraction, respectively. 



Inspired by traditional information extraction tasks~\cite{wang2025fintaggingllmreadybenchmarkextracting,wang2022conditional}, \textsc{EPPCMinerBen} follows a hierarchical modeling structure where the first task assigns high-level codes to each sentence. Given these codes, the second task selects the most appropriate subcode from each code-specific set, enabling hierarchical, coarse-to-fine interpretation. The final task extracts minimal text spans that justify each code–subcode pair. Together, these tasks provide a structured framework for modeling the semantic and relational dynamics in patient-provider communication. The hierarchical design enables EPPCMinerBen to evaluate discourse understanding, hierarchical reasoning, semantic precision, and grounding-based interpretability, which are capabilities that traditional flat label benchmarks cannot effectively measure.

\subsubsection{EPPC Code Classification}\label{subsubsec2}

The EPPC code classification is a sentence-level multi-label classification task that evaluates the coarse-grained semantic understanding capability of LLMs. Specifically, given a current sentence $s$ along with its context $C$, and a predefined set of code categories $\mathcal{L}_{code}=\{l_1,l_2,\cdots,l_9\}$, the LLM is tasked with generating a set of relevant code labels for $s$.

\begin{equation}
    f:(s,C)\rightarrow\hat{Y}_{code}\subseteq\mathcal{L}_{code}
\end{equation}
where $\hat{Y}_{code}$ is the predicted set of code labels. The function $f$ is implemented by LLMs prompted to perform multi-label classification at the sentence level.

\subsubsection{EPPC Subcode Classification}
The EPPC sub-code classification is a conditional multi-class classification task that refines the interpretation of each predicted high-level code by selecting the most appropriate subcode from a predefined, code-specific candidate set. Specifically, given the current sentence $s$, its context $C$, the predicted code set $\hat{Y}_{code}=\{l_1,l_2,\cdots,l_m\}$ where $m \leq 9$, a predefined mapping from each code $l_j$ to its corresponding subcode set $\mathcal{L}^{(l_j)}_{sub}=\{t^{(j)}_1,t^{(j)}_2,\cdots,t^{(j)}_{k_j}\}$, the task aims to select the most appropriate subcode for each $l_j$ to form a valid code-subcode pair.

\begin{equation}
    f:(s,C,\hat{Y}_{code})\rightarrow\hat{Y}_{sub}
\end{equation}
where $\hat{Y}_{sub}=\{(l_j,\hat{t}^{(j)})|\hat{t}^{(j)}\in \mathcal{L}_{sub}^{(l_j)},l_j\in \hat{Y}_{code} \}$ is the formed code-subcode pair, $\hat{t}^{(j)}$ is the predicted subcode for code $l_j$.

\subsubsection{EPPC Evidence Extraction}
The EPPC evidence extraction task aims to identify the minimal span within the sentence that supports each predicted code-subcode pair. It focuses on extracting concise textual evidence that justifies the assigned labels. Specifically, given the current sentence $s$, its context $C$, and code-subcode pairs $\hat{Y}_{sub}$, the task is to extract a corresponding evidence span $e^{(j)}\subseteq s$ for each pair that reflects the semantic rationale behind the prediction. Therefore, the evidence extraction function is defined as:

\begin{equation}
    f:(s,C,\hat{Y}_{sub})\rightarrow \hat{Y}_{evid}
\end{equation}
where $\hat{Y}_{evid}=\{(l_j,\hat{t}^{(j)},e^{(j)})|e^{(j)}\subseteq s \}$ is the final structured triplet, $e^{(j)}$ is the minimal span (a contiguous or discontinuous sequence of tokens) within $s$ that provides support for the pair $(l_j,\hat{t}^{(j)})$. Thus, $(code_j,subcode_j,evidence_j) \equiv (l_j,\hat{t}^{(j)},e^{(j)})$.

\subsection{Benchmark Data Annotation}
\subsubsection{Data description}
As shown in Table~\ref{tab:message_statistics}, we collected 752 secure messages (SMs) from Yale New Haven Health (YNHH), comprising 449 patient and 303 provider messages. These messages, annotated with EPPC codes and phrases, will be released via the Cancer Data Service (CDS), part of the NCI’s Cancer Research Data Commons. The messages were segmented into 1,933 unique sentences. Using NLTK and the cl100k\_base tokenizer, we obtained 27,849 words and 33,388 tokens in total. On average, each message contains 37.03 words (SD=33.35) and 44.40 tokens (SD=40.01), with the longest message reaching 248 words or 362 tokens.


\begin{table}[!t]
\scriptsize
\caption{The statistics of the collected secure message.\label{tab:message_statistics}}
\centering
\begin{tabular}{p{0.5\columnwidth}l}
\hline
\textbf{Item} & \textbf{Information} \\
\hline
Data Source & YNHH \\
\# Total Messages & 752 \\
\# Patient’s Message & 449 \\
\# Provider’s Message & 303 \\
\# Sentences & 1,933 \\
\# Total words & 27,849 \\
Ave. Words $\pm$ std & 37.03 $\pm$ 33.35 \\
Max Words & 248 \\
\# Total tokens & 33,388 \\
Ave. Tokens $\pm$ std & 44.40 $\pm$ 40.01 \\
Max Tokens & 362 \\
\hline
\end{tabular}
\end{table}


\subsubsection{Annotation}
\begin{figure}[ht]
\centering

\subfigure[\scriptsize  The distribution of EPPC Codes.\label{fig:eppc_codes}]{
    \includegraphics[width=0.6\linewidth]{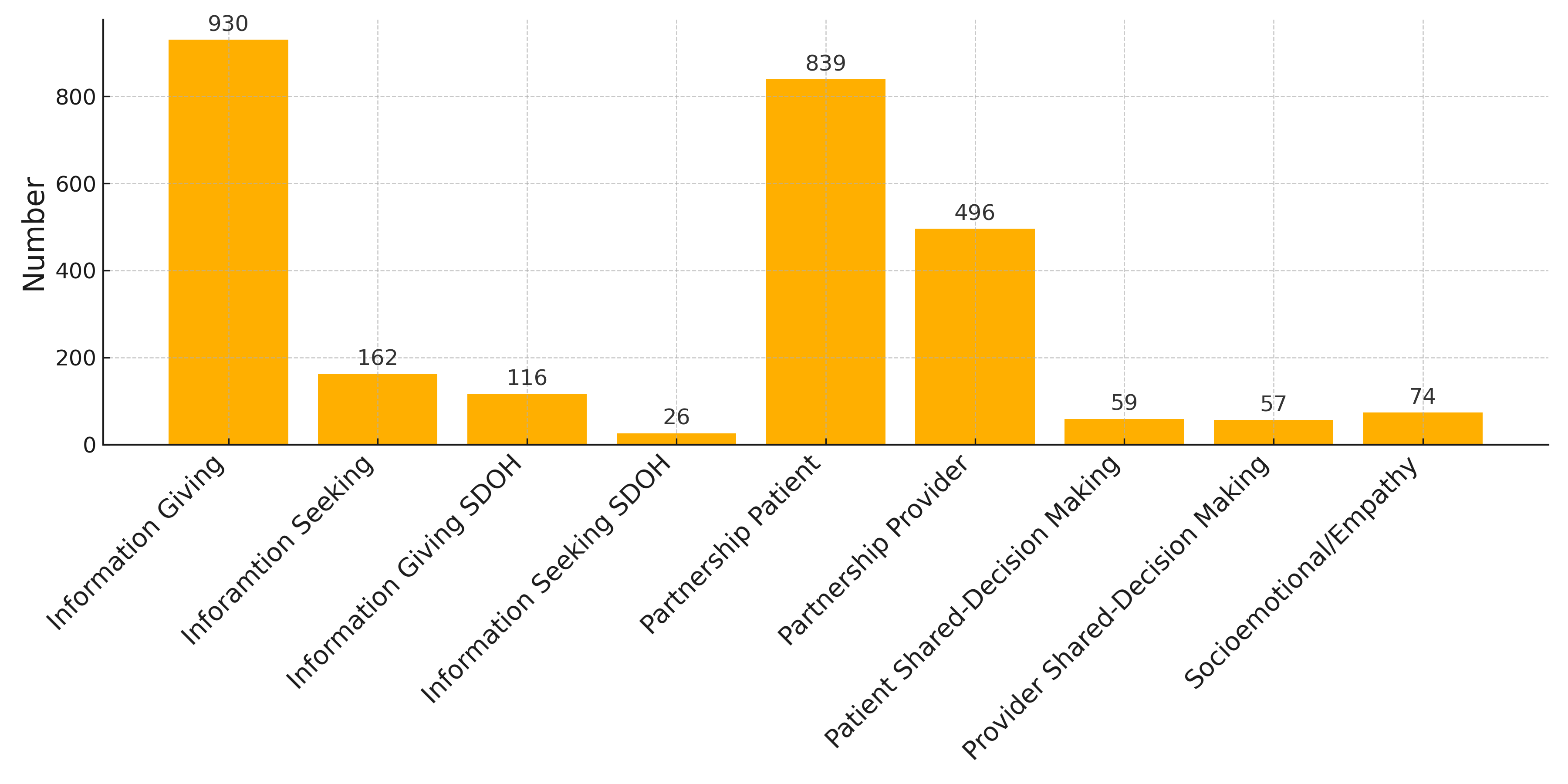}
}

\subfigure[\scriptsize  The distribution of EPPC subcodes.\label{fig:eppc_subcodes}]{
    \includegraphics[width=0.6\linewidth]{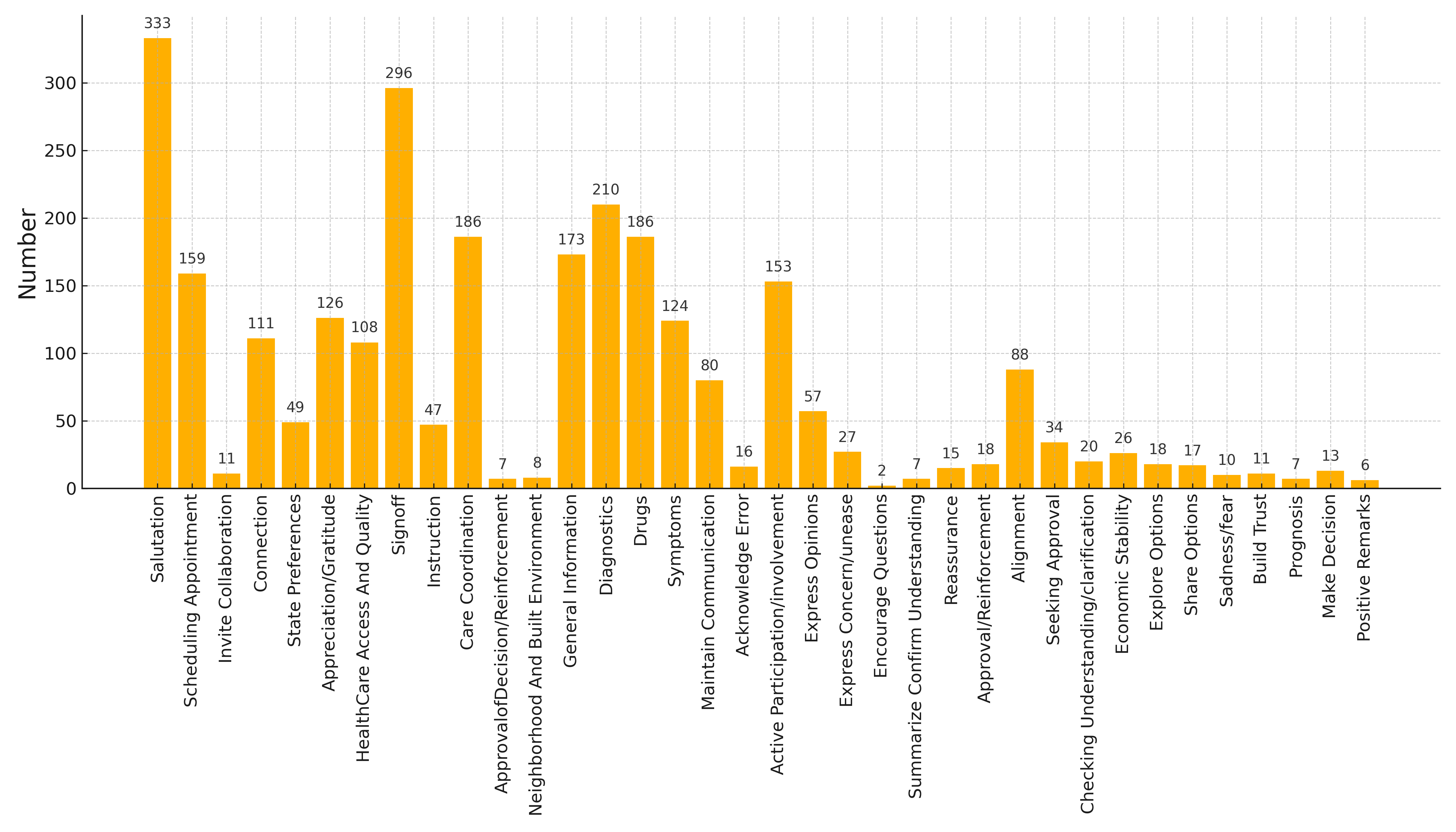}
}

\caption{The distribution of EPPC codes and subcodes based on the annotation.}
\label{fig:eppc_codes_and_subcodes}
\end{figure}

Based on 752 SM threads, annotation was conducted using a structured EPPC codebook inspired by the Roter interaction analysis system (RIAS)  manual \cite {roter2002roter} for coding medical dialogue. EPPC codebook covers 9 communication behaviors (codes) and their corresponding subcodes (Figure~\ref{fig_1}). The distribution of codes and subcodes is shown in Figure~\ref{fig:eppc_codes_and_subcodes}. “Information Giving” (930) and “Patient Partnership” (839) are the most frequent codes, emphasizing information exchange and patient engagement. At the subcode level, common types include “Salutation” (333), “Signoff” (296), “Diagnostics” (210), and “Drugs” (186), while emotionally supportive cues like “Encourage Questions” (2) and “Positive Remarks” (6) are rare, indicating limited affective communication in secure messaging.

The annotation process began with the full research team collaboratively labeled the first 100 messages in samples of 20 during multiple meetings, which ensured a shared understanding of the EPPC codes, consistent application across annotators, and early refinement of the EPPC codebook. After this initial calibration phase, which established shared knowledge and proficiency in annotating communication patterns, two members of the research team independently labeled each secure message thread \footnote{A thread refers to a sequence of back-and-forth communication between a patient and their healthcare providers, typically centered around a specific concern, topic, or episode of care.} using the EPPC codebook. A total of 13 samples of 50 messages were manually annotated. For example, the phrase "It is not showing up on my appointment list " was annotated with the Appointment Related Information subcode under the Information Giving Code, "Do I need to do anything else?" was coded as Active Participation and Involvement subcode under the partnership code on the patient side, and "Can you please update my Medication list" was coded as Seeking Approval subcode under Shared-Decision Making code on the patient side.  The Agreement scores were tracked per sample and improved over time. The average agreement score between the annotators  evaluated using Cohen's Kappa was 0.74 (sample 1=.67, median sample=.74, sample 13 =.76). Discrepancies were resolved by an adjudicator and through consensus meetings led by senior investigators with experience in communication coding and qualitative methods, which also helped refine communication codes definitions and enhance the calibration of the annotator. 

Beyond consensus resolution, a senior reviewer performed random audits and spot-checks to ensure label accuracy and adherence to the codebook. Ambiguous cases and emerging linguistic patterns were discussed collaboratively, with feedback incorporated into the evolving annotation manual. This multi-level quality assurance process resulted in a high-fidelity annotated dataset, providing a robust foundation for training and evaluating models in \textsc{EPPCMinerBen}.

\subsection{Prompt Engineering}
To evaluate large language models (LLMs) on the \textsc{EPPCMinerBen} task, we adopt both zero-shot and few-shot prompting strategies. In zero-shot prompting, the model is given only a task description without any examples, relying entirely on its prior knowledge to generate responses. While this approach showcases the model’s general capabilities, it can struggle with complex or nuanced tasks due to the absence of contextual guidance. Few-shot prompting addresses this limitation by incorporating a small number of in-context examples within the prompt, which help illustrate the desired output format and task expectations. These demonstrations serve as learning cues that improve the model’s ability to generalize, enabling more accurate and contextually appropriate predictions in challenging scenarios. Figure~\ref{fig:prompt-template}
 in Appendix A illustrates a prompt template. 

\subsection{LLMs Evaluation}
To benchmark performance, we evaluate models across different parameter scales, including large-scale open-source models, medium-scale open-source models, smaller-scale open-source models, and clinical-domain models. Specifically, for large-scale open-source models, we report Llama-3.3-70b-Instruct~\cite{Dubey2024TheL3}, Llama-3.1-70b-Instruct~\cite{Dubey2024TheL3}, and DeepSeek-R1-Distill-Llama-70b~\cite{guo2025deepseek}. For medium-scale models, we include QwQ-32b-AWQ~\cite{qwen2025qwen25technicalreport,qwq32b}, Gemma-2-27b-it~\cite{team2024gemma,team2024gemma1}, DeepSeek-R1-Distill-Qwen-32b~\cite{guo2025deepseek} and Qwen2.5-14b-Instruct-1M~\cite{yang2025qwen2}. For small-scale models, we report Llama-3.1-8b-Instruct~\cite{Dubey2024TheL3}, Llama-3.2-3b-Instruct~\cite{Dubey2024TheL3}, Qwen2.5-1.5b-Instruc~\cite{qwen2025qwen25technicalreport,qwen2.5} and Llama-3.2-1b-Instruct~\cite{Dubey2024TheL3}. For clinical-domain models, we introduce Llama3-OpenBioLLM-70B~\cite{OpenBioLLMs} and sdoh-llama-3-3-70b~\cite{landes2025integration} models. We included the SDOH model specifically because we annotated SDOH factors, such as
housing instability, food insecurity, and lack of social support and these factors are often found in free
text records and are considered vital to understanding health disparities and providing
personalized care. We present statistical information in Table~\ref{tab:llm_statistics}.

\begin{table}[t]
\scriptsize
\caption{Statistics for evaluated large language models.\label{tab:llm_statistics}}
\centering
\resizebox{\columnwidth}{!}{%
\begin{tabular}{lllc}
\hline
\textbf{Category} & \textbf{Model} & \textbf{Size} & \textbf{Source} \\
\hline
\multirow{3}{*}{Large-scale open-source LLMs} 
    & Llama-3.3-70B-Instruct & 70B & meta-llama/Llama-3.3-70B-Instruct \\
    & Llama-3.1-70B-Instruct & 70B & meta-llama/Llama-3.1-70B-Instruct \\
    & DeepSeek-R1-Distill-Llama-70B & 70B & deepseek-ai/DeepSeek-R1-Distill-Llama-70B \\
\hline
\multirow{4}{*}{Medium-scale open-source LLMs}
    & QwQ-32B-AWQ & 32B & Qwen/QwQ-32B-AWQ \\
    & Gemma-2-27B-it & 27B & google/gemma-2-27b-it \\
    & DeepSeek-R1-Distill-Qwen-32B & 32B & deepseek-ai/DeepSeek-R1-Distill-Qwen-32B \\
    & Qwen2.5-14B-Instruct-1M & 14B & Qwen/Qwen2.5-14B-Instruct-1M \\
\hline
\multirow{5}{*}{Small-scale open-source LLMs}
    & Llama-3.1-8B-Instruct & 8B & meta-llama/Llama-3.1-8B-Instruct \\
    & Llama-3.2-3B-Instruct & 3B & meta-llama/Llama-3.2-3B-Instruct \\
    & Qwen2.5-1.5B-Instruct & 1.5B & Qwen/Qwen2.5-1.5B-Instruct \\
    & Llama-3.2-1B-Instruct & 1B & meta-llama/Llama-3.2-1B-Instruct \\
\hline
\multirow{3}{*}{Clinical-domain LLMs}
    & Llama3-OpenBioLLM-70B & 70B & aaditya/Llama3-OpenBioLLM-70B \\
    & sdoh-llama-3-3-70B & 70B & plandes/sdoh-llama-3-3-70b \\
\hline
\end{tabular}%
}
\end{table}

\subsection{Evaluation Metrics}
We evaluate model performance across the three \textsc{EPPCMinerBen} subtasks using micro-averaged Precision, Recall, and F1-score. Code and subcode classification are treated as multi-label classification tasks, with metrics computed by comparing predicted and gold label sets at the message level. For evidence extraction, we adopt a relaxed token-level matching strategy that considers both span containment and Jaccard similarity (threshold $\geq$ 0.6) to account for partial overlaps. 

\section{Results}

\subsection{EPPC Code Classification Performance}
Table~\ref{tab:eppc_code_performance} presents the performance of various large language models (LLMs) on the EPPC code classification task under 0-shot, 1-shot, and 2-shot settings. The results highlight the dominance of large-scale instruction-tuned models. {LLaMA-3.3-70B-Instruct} achieves the highest overall performance, peaking at \textbf{67.03\%} F1 in the 2-shot setting, demonstrating effective in-context learning capabilities. Similarly, {LLaMA-3.1-70B-Instruct} maintains consistent high performance across all settings. Interestingly, the distilled model {DeepSeek-R1-Distill-Qwen-32B} shows remarkable scalability with few-shot prompting, improving from 61.52\% (0-shot) to 66.45\% (2-shot), rivaling the performance of significantly larger 70B parameter models.

In contrast, mid-sized models exhibit divergent behaviors. {Gemma-2-27b-it} delivers exceptional 0-shot performance (64.62\%), comparable to the best 70B models, but suffers a significant performance degradation in few-shot settings, suggesting a sensitivity to prompt context or "context overflow" where examples confuse rather than aid the model. {QwQ-32} and {Qwen2.5-14B-Instruct-1M} show more stable, albeit modest, improvements or plateaus with additional shots.

For smaller models, the impact of few-shot prompting is critical. {Llama-3.1-8B-Instruct} completely fails in the 0-shot setting (0.00\%), likely due to an inability to follow the strict output format without examples, but recovers substantially to over 51\% with just one example. Lightweight models (under 3B parameters) generally underperform; however, {Qwen2.5-1.5B-Instruct} shows a notable jump from 0-shot to 1-shot, whereas the {Llama-3.2} series (1B and 3B) struggles to achieve competitive scores even with prompting.

Regarding domain-specific models, {sdoh-llama-3-3-70b} demonstrates robust performance (up to 64.68\%), closely tracking its base model, indicating that domain adaptation did not compromise its general reasoning abilities for this task. Conversely, {Llama3-OpenBioLLM-70B} lags behind general-purpose instructors, underscoring that medical domain pre-training alone does not guarantee superior performance on specific discourse classification tasks without strong instruction alignment.

\begin{table*}[t]
\scriptsize
\centering
\setlength{\tabcolsep}{6pt}
\caption{F1 performance (\%) for EPPC code classification under different shot settings.}\label{tab:eppc_code_performance}
\begin{tabular}{lccc}
\toprule
\textbf{Model} & \textbf{0 shot F1} & \textbf{1 shot F1} & \textbf{2 shots F1} \\
\midrule
LLaMA-3.3-70B-Instruct 
& 64.21 ± 0.00e+00 
& 64.60 ± 1.56e-01 
& 67.03 ± 1.00e-01 \\

LLaMA-3.1-70B-Instruct
& 64.57 ± 9.20e-02
& 64.93 ± 8.00e-02
& 65.40 ± 1.20e-02 \\

DeepSeek-R1-Distill-Llama-70B
& 59.43 ± 2.43e-01
& 58.58 ± 0.00e+00
& 61.89 ± 6.40e-02 \\

Gemma-2-27b-it
& 64.62 ± 0.00e+00
& 53.22 ± 0.00e+00
& 52.67 ± 0.00e+00 \\

DeepSeek-R1-Distill-Qwen-32B
& 61.52 ± 0.00e+00
& 64.62 ± 0.00e+00
& 66.45 ± 0.00e+00 \\

QwQ-32
& 60.19 ± 0.00e+00
& 61.97 ± 0.00e+00
& 63.36 ± 0.00e+00 \\

Qwen2.5-14B-Instruct-1M
& 60.63 ± 2.10e-01
& 60.92 ± 1.66e-01
& 60.58 ± 6.79e-01 \\

Llama-3.1-8B-Instruct
& 0.00 ± 0.00e+00
& 51.08 ± 0.00e+00
& 51.08 ± 0.00e+00 \\

Llama-3.2-3B-Instruct
& 30.30 ± 2.42e-01
& 39.77 ± 0.00e+00
& 39.77 ± 0.00e+00 \\

Llama-3.2-1B-Instruct
& 27.16 ± 9.70e-02
& 24.96 ± 0.00e+00
& 24.96 ± 0.00e+00 \\

Qwen2.5-1.5B-Instruct
& 25.88 ± 3.80e-02
& 38.05 ± 0.00e+00
& 38.05 ± 0.00e+00 \\

Llama3-OpenBioLLM-70B
& 51.46 ± 0.00e+00
& 52.76 ± 0.00e+00
& 52.62 ± 0.00e+00 \\

sdoh-llama-3-3-70b
& 63.60 ± 0.00e+00
& 63.85 ± 0.00e+00
& 64.68 ± 0.00e+00 \\

\bottomrule
\end{tabular}
\end{table*}

\subsection{EPPC Subcode Classification Performance}
Table~\ref{tab:eppc_subcode_performance} presents the performance of various LLMs on the EPPC subcode classification task across 0-shot, 1-shot, and 2-shot settings. Compared to the coarse-grained code classification, all models exhibit a substantial drop in F1 scores, with the best models hovering around 48\%, underscoring the increased difficulty of distinguishing fine-grained communicative intentions.

The results reveal that model size is not the sole determinant of performance in this nuanced task. The distillation-based models, particularly {DeepSeek-R1-Distill-Qwen-32B} and {DeepSeek-R1-Distill-Llama-70B}, deliver the strongest results. Notably, the 32B parameter Qwen-distilled model achieves the highest 0-shot performance (\textbf{48.25\%}) and maintains robust stability across shot settings, outperforming the larger {LLaMA-3.3-70B-Instruct}, which peaks at 44.46\% in the 2-shot setting. This suggests that the reasoning capabilities distilled into the DeepSeek series are particularly beneficial for hierarchical classification.

QwQ-32 also performs impressively, starting at 45.35\% and improving to 46.98\% with two shots, further indicating that models optimized for reasoning (Chain-of-Thought style or similar) handle subcode ambiguity better than standard instruction-tuned models. In contrast, {Gemma-2-27b-it} repeats its behavior from the coarse-grained task: decent 0-shot capability (42.13\%) followed by a sharp performance collapse in few-shot settings (dropping to $\sim$30\%), confirming a significant alignment issue with in-context examples.

For smaller models, the task proves overwhelming. {Llama-3.1-8B-Instruct} again fails in the 0-shot setting (0.00\%) due to formatting errors but recovers to a modest $\sim$24\% with examples. Models under 3B parameters generally fail to extract meaningful signals, with F1 scores staying below 16\%. Among domain-specific models, {sdoh-llama-3-3-70b} (40.80\%--42.50\%) performs comparably to its base model but does not offer a distinct advantage, while {Llama3-OpenBioLLM-70B} significantly underperforms ($<30\%$), highlighting that broad biomedical knowledge does not automatically translate to the specific socio-pragmatic reasoning required for EPPC subcodes.

\begin{table*}[t]
\scriptsize
\centering
\setlength{\tabcolsep}{6pt}
\caption{F1 performance (\%) for EPPC subcode classification under different shot settings.}\label{tab:eppc_subcode_performance}
\begin{tabular}{lccc}
\toprule
\textbf{Model} & \textbf{0 shot F1} & \textbf{1 shot F1} & \textbf{2 shots F1} \\
\midrule

LLaMA-3.3-70B-Instruct
& 41.70 $\pm$ 0.00e+00
& 42.19 $\pm$ 1.16e-01
& 44.46 $\pm$ 8.40e-02 \\

LLaMA-3.1-70B-Instruct
& 43.08 $\pm$ 6.80e-02
& 44.22 $\pm$ 4.00e-02
& 44.15 $\pm$ 2.00e-02 \\

DeepSeek-R1-Distill-Llama-70B
& 48.17 $\pm$ 1.52e-01
& 44.92 $\pm$ 0.00e+00
& 46.15 $\pm$ 3.60e-02 \\

Gemma-2-27b-it
& 42.13 $\pm$ 0.00e+00
& 31.82 $\pm$ 0.00e+00
& 30.88 $\pm$ 0.00e+00 \\

DeepSeek-R1-Distill-Qwen-32B
& 48.25 $\pm$ 5.55e-15
& 47.87 $\pm$ 0.00e+00
& 47.95 $\pm$ 0.00e+00 \\

QwQ-32
& 45.35 $\pm$ 0.00e+00
& 45.94 $\pm$ 5.55e-15
& 46.98 $\pm$ 5.55e-15 \\

Qwen2.5-14B-Instruct-1M
& 42.55 $\pm$ 7.05e-02
& 41.83 $\pm$ 1.53e-01
& 41.56 $\pm$ 8.74e-01 \\

Llama-3.1-8B-Instruct
& 0.00 $\pm$ 0.00e+00
& 24.66 $\pm$ 0.00e+00
& 24.66 $\pm$ 0.00e+00 \\

Llama-3.2-3B-Instruct
& 12.31 $\pm$ 4.58e-02
& 15.65 $\pm$ 0.00e+00
& 15.65 $\pm$ 0.00e+00 \\

Llama-3.2-1B-Instruct
& 5.26 $\pm$ 6.97e-02
& 8.26 $\pm$ 0.00e+00
& 8.26 $\pm$ 0.00e+00 \\

Qwen2.5-1.5B-Instruct
& 7.24 $\pm$ 1.09e-01
& 14.47 $\pm$ 0.00e+00
& 14.47 $\pm$ 0.00e+00 \\

Llama3-OpenBioLLM-70B
& 29.03 $\pm$ 0.00e+00
& 27.11 $\pm$ 0.00e+00
& 26.88 $\pm$ 0.00e+00 \\

sdoh-llama-3-3-70b
& 40.80 $\pm$ 5.55e-15
& 41.27 $\pm$ 5.55e-15
& 42.50 $\pm$ 0.00e+00 \\

\bottomrule
\end{tabular}
\end{table*}

\subsection{EPPC Evidence Extraction Performance}
{Table~\ref{tab:eppc_evidence_performance} reports model performance on the EPPC evidence extraction task, measured using Jaccard similarity between predicted and gold evidence spans. Unlike the classification tasks, where models must abstractly label intent, this task requires precise token-level grounding to extract the specific linguistic phrases supporting the code.

The results indicate that evidence extraction is comparatively easier for high-capacity models than subcode classification, with top models achieving significantly higher F1 scores. {LLaMA-3.1-70B-Instruct} emerges as the clear state-of-the-art, demonstrating exceptional stability and peaking at \textbf{82.84\%} in the 2-shot setting. Interestingly, it outperforms the newer {LLaMA-3.3-70B-Instruct}, which exhibits volatility, dropping from 71.01\% (0-shot) to 66.88\% (1-shot) before recovering to 78.84\% (2-shot).

Among mid-sized and distilled models, {DeepSeek-R1-Distill-Qwen-32B} (76.12\% 0-shot) continues to impress, significantly outperforming the larger {DeepSeek-R1-Distill-Llama-70B} and rivaling the top-tier Llama models. {Qwen2.5-14B-Instruct} also performs surprisingly well in the 0-shot setting (76.69\%), indicating that current instruction-tuned models are inherently effective at “copy-paste” style extraction without requiring many examples.

A notable anomaly is observed in {Llama-3.1-8B-Instruct}. Unlike the classification tasks where few-shot prompting rescued its performance, here, the model performs best in 0-shot (41.15\%) and degrades with examples ($\sim$36\%). This suggests that for extraction tasks, the additional context in prompts might introduce noise or formatting confusion for smaller models. Similarly, {Gemma-2-27b-it} and domain-specific models like {sdoh-llama-3-3-70b} exhibit a "negative transfer" effect, where performance slighty declines or stagnates as shots are added.

Conversely, {Qwen2.5-1.5B-Instruct} is the only small model that benefits substantially from prompting, jumping from 36.17\% to 50.10\%. Overall, while the task yields higher absolute scores than classification, it reveals a distinct "less is more" trend for several models regarding prompt context.

\begin{table*}[t]
\scriptsize
\centering
\setlength{\tabcolsep}{6pt}
\caption{F1 performance (\%) based on Jaccard similarity for EPPC evidence extraction under different shot settings.}\label{tab:eppc_evidence_performance}
\begin{tabular}{lccc}
\toprule
\textbf{Model} & \textbf{0 shot F1} & \textbf{1 shot F1} & \textbf{2 shots F1} \\
\midrule

LLaMA-3.3-70B-Instruct
& 71.01 $\pm$ 0.00e+00
& 66.88 $\pm$ 4.20e-01
& 78.84 $\pm$ 7.52e-01 \\

LLaMA-3.1-70B-Instruct
& 78.87 $\pm$ 7.60e-02
& 82.06 $\pm$ 2.00e-02
& 82.84 $\pm$ 6.00e-02 \\

DeepSeek-R1-Distill-Llama-70B
& 72.98 $\pm$ 1.08e-01
& 72.45 $\pm$ 0.00e+00
& 73.90 $\pm$ 1.28e-01 \\

Gemma-2-27b-it
& 69.92 $\pm$ 0.00e+00
& 67.14 $\pm$ 0.00e+00
& 67.31 $\pm$ 0.00e+00 \\

DeepSeek-R1-Distill-Qwen-32B
& 76.12 $\pm$ 0.00e+00
& 75.35 $\pm$ 0.00e+00
& 77.24 $\pm$ 0.00e+00 \\

QwQ-32
& 65.11 $\pm$ 0.00e+00
& 64.83 $\pm$ 0.00e+00
& 66.14 $\pm$ 0.00e+00 \\

Qwen2.5-14B-Instruct-1M
& 76.69 $\pm$ 1.67e-01
& 73.68 $\pm$ 1.35e-01
& 74.00 $\pm$ 6.42e-01 \\

Llama-3.1-8B-Instruct
& 41.15 $\pm$ 5.55e-15
& 36.42 $\pm$ 0.00e+00
& 36.42 $\pm$ 0.00e+00 \\

Llama-3.2-3B-Instruct
& 15.25 $\pm$ 1.34e-01
& 32.47 $\pm$ 0.00e+00
& 32.47 $\pm$ 0.00e+00 \\

Llama-3.2-1B-Instruct
& 8.86 $\pm$ 1.86e-02
& 6.84 $\pm$ 0.00e+00
& 6.84 $\pm$ 0.00e+00 \\

Qwen2.5-1.5B-Instruct
& 36.17 $\pm$ 1.88e-01
& 50.10 $\pm$ 0.00e+00
& 50.10 $\pm$ 0.00e+00 \\

Llama3-OpenBioLLM-70B
& 70.65 $\pm$ 0.00e+00
& 69.89 $\pm$ 0.00e+00
& 69.18 $\pm$ 0.00e+00 \\

sdoh-llama-3-3-70b
& 73.42 $\pm$ 0.00e+00
& 70.16 $\pm$ 0.00e+00
& 70.05 $\pm$ 0.00e+00 \\

\bottomrule
\end{tabular}
\end{table*}

\subsection{Statistical Significance Analysis }\label{sec5}

To ensure statistical robustness, every model–task–shot configuration was evaluated with multiple independent runs, each using different random seeds, and we report the resulting mean $\pm$ standard deviation in scientific notation. This presentation provides a clearer view of performance stability, model sensitivity to randomness, and the reliability of model comparisons. We further verified that the relative model rankings remain consistent under these repeated runs.

In addition, a brief analysis of the reported variances shows that larger models exhibit extremely low standard deviations across runs, indicating highly stable behavior on all three EPPC tasks. Mid-size models display moderate variability, especially on subcode and Evidence Extraction, suggesting that these tasks are more sensitive to decision boundary fluctuations. Smaller models show higher variance and occasionally degraded predictions (e.g., extremely low recall), highlighting their instability in more complex EPPC subtasks. These patterns further support the need for multi-run reporting and reinforce the reliability of our performance comparisons.

\begin{figure}
    \centering
    \includegraphics[width=1\linewidth]{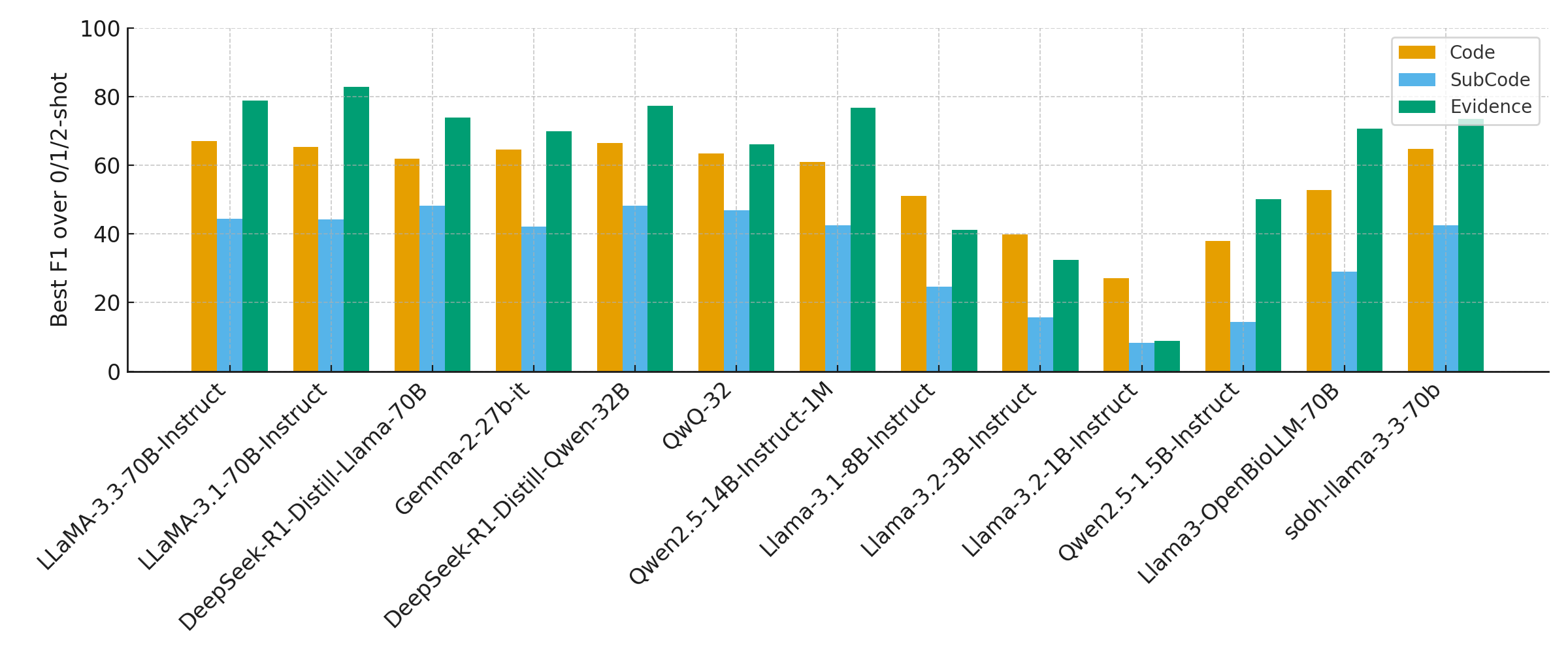}
    \caption{EPPC overall F1 results across all EPPC subtasks}
    \label{fig:overall_preformance}
\end{figure}

{Figure~\ref{fig:overall_preformance} provides a global comparison of model performance across all EPPC subtasks. For each model, we plot the best F1 score obtained over the 0-shot, 1-shot, and 2-shot settings on code, subcode, and evidence extraction. Larger models (e.g., 70B variants) consistently dominate across subtasks, while smaller 1–3B models struggle especially on the more fine-grained SubCode and Evidence tasks, highlighting the increased difficulty of modeling relational and span-level EPPC phenomena.

We acknowledge that several potential confounding factors may have influenced the observed performance differences across models. In particular, variations in prompt design, differences in domain familiarity, and potential mismatches in pretraining data could contribute to these trends alongside model size or instruction tuning.

Our results show that large, instruction-tuned models tend to perform better in \textsc{EPPCMinerBen} tasks, especially evidence extraction, a trend that may be associated with their scale and alignment. Smaller models, on the other hand, struggle with fine-grained reasoning.

\section{Discussion}

\subsection{EPPCMinerBen Tasks Comparative Analysis}

Figure~\ref{fig:major_anay} and Tables~\ref{tab:eppc_code_performance}–\ref{tab:eppc_evidence_performance} collectively highlight key trends across the three \textsc{EPPCMinerBen} subtasks, offering insight into model behavior and reasoning performance in clinical communication.

\begin{figure}
    \centering
    \includegraphics[width=1\linewidth]{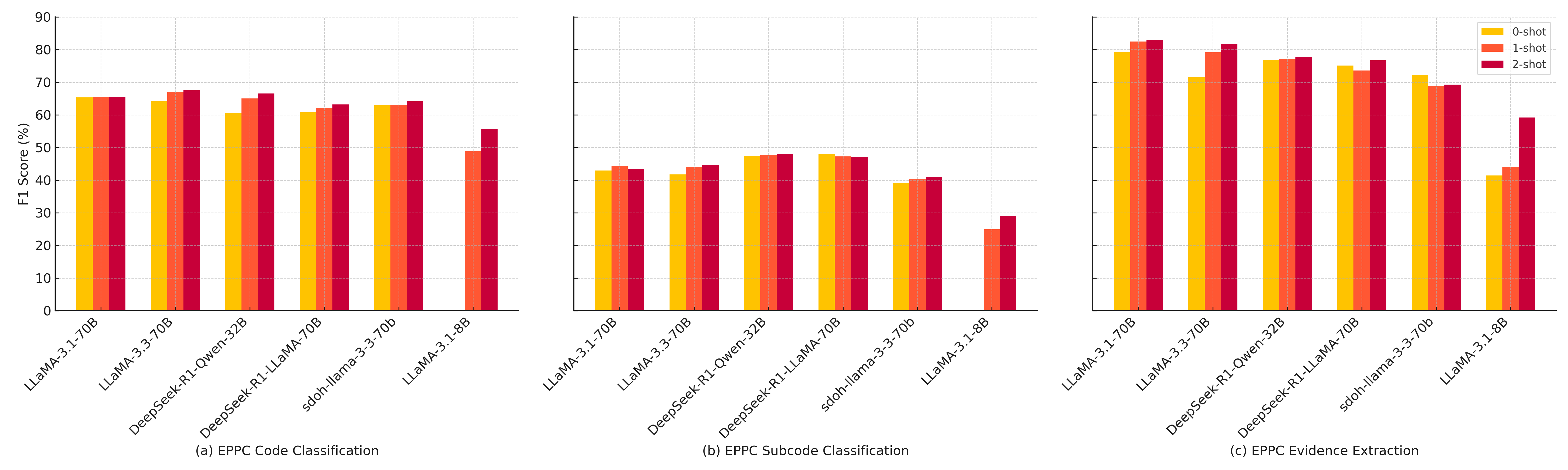}
    \caption{Cross-task performance comparison based on \textsc{EPPCMinerBen} results}
    \label{fig:major_anay}
\end{figure}

\textsc{EPPCMinerBen} reveals critical insights into model capabilities and limitations across discourse-level clinical NLP tasks. First, instruction tuning and model scale jointly drive generalization. Large, instruction-aligned models such as {Llama-3.1-70B} and {DeepSeek-R1-Distill} consistently outperform smaller ones, confirming that strong task comprehension outweighs parameter count. Distilled models often match full-scale LLMs, highlighting the benchmark’s ability to assess efficiency-performance trade-offs. Second, few-shot prompting improves results, especially in span-level tasks, but its effectiveness depends heavily on model design and prompt quality. This demonstrates the benchmark's sensitivity to prompt construction and calibration. Third, subcode classification is the hardest task, requiring fine-grained, hierarchical reasoning. It exposes models’ struggles to disambiguate similar labels, making the benchmark an effective probe for deep semantic understanding under structural constraints. Fourth, evidence extraction shows the largest gap between strong and weak models, indicating that only well-aligned models can achieve accurate span-level grounding. This makes the task a strong indicator of contextual alignment and output interpretability.

\begin{figure*}[t]
\centering

\begin{minipage}[t]{0.32\textwidth}
    \centering
    \includegraphics[width=\linewidth]{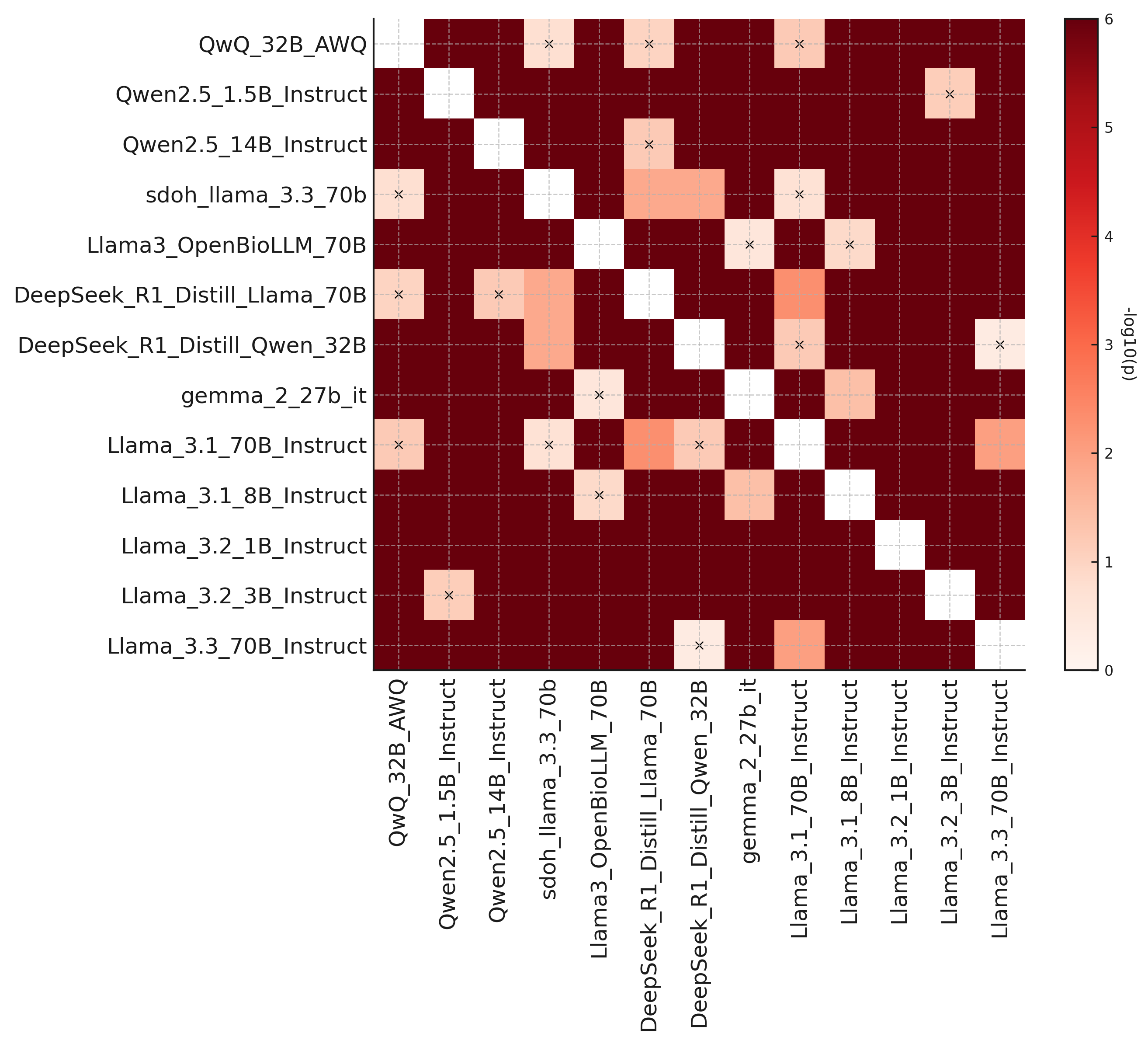}
    \caption*{(a) bootstrap test based on Code F1}
\end{minipage}
\hfill
\begin{minipage}[t]{0.32\textwidth}
    \centering
    \includegraphics[width=\linewidth]{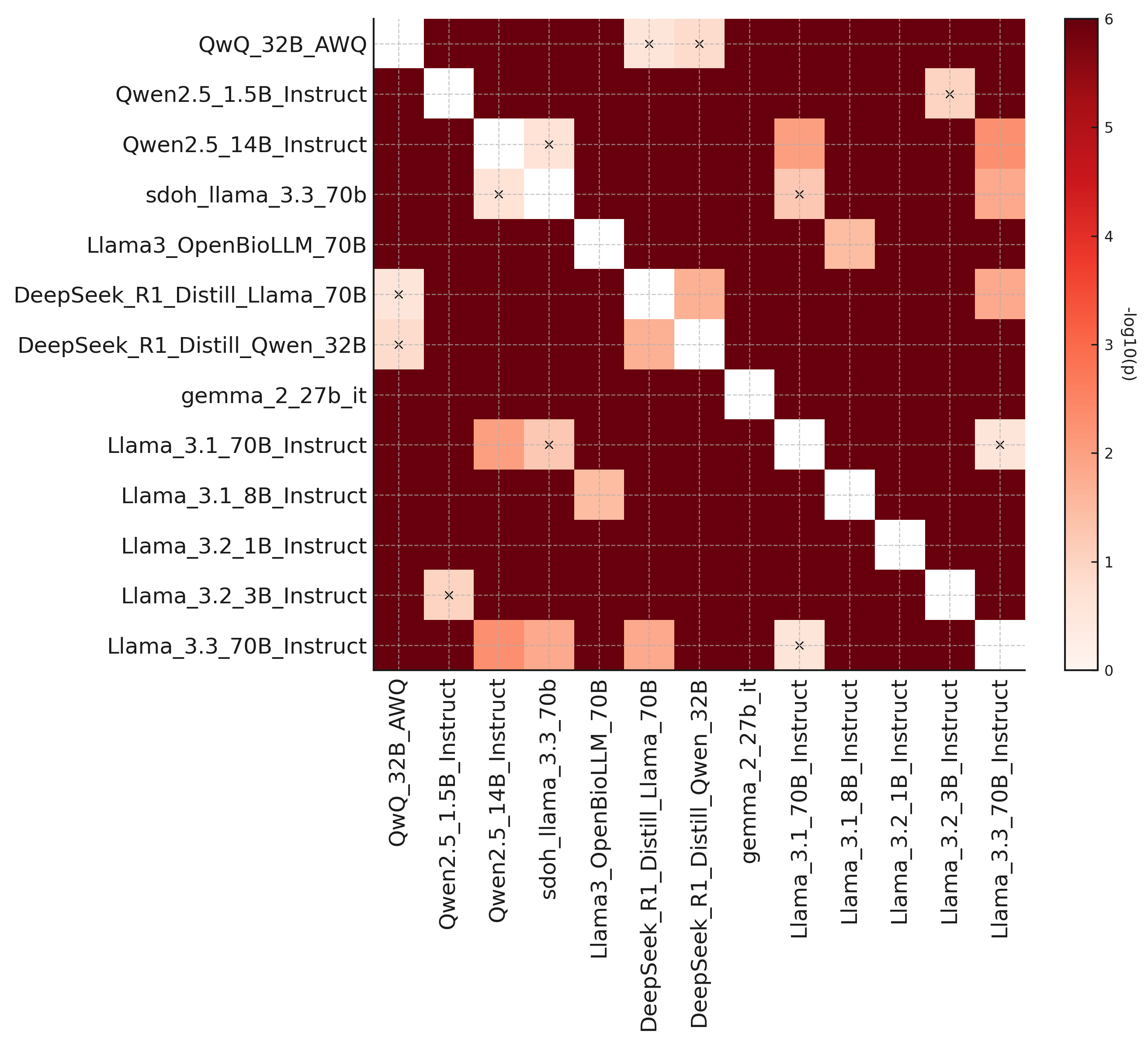}
    \caption*{(b) bootstrap test based on SubCode F1}
\end{minipage}
\hfill
\begin{minipage}[t]{0.32\textwidth}
    \centering
    \includegraphics[width=\linewidth]{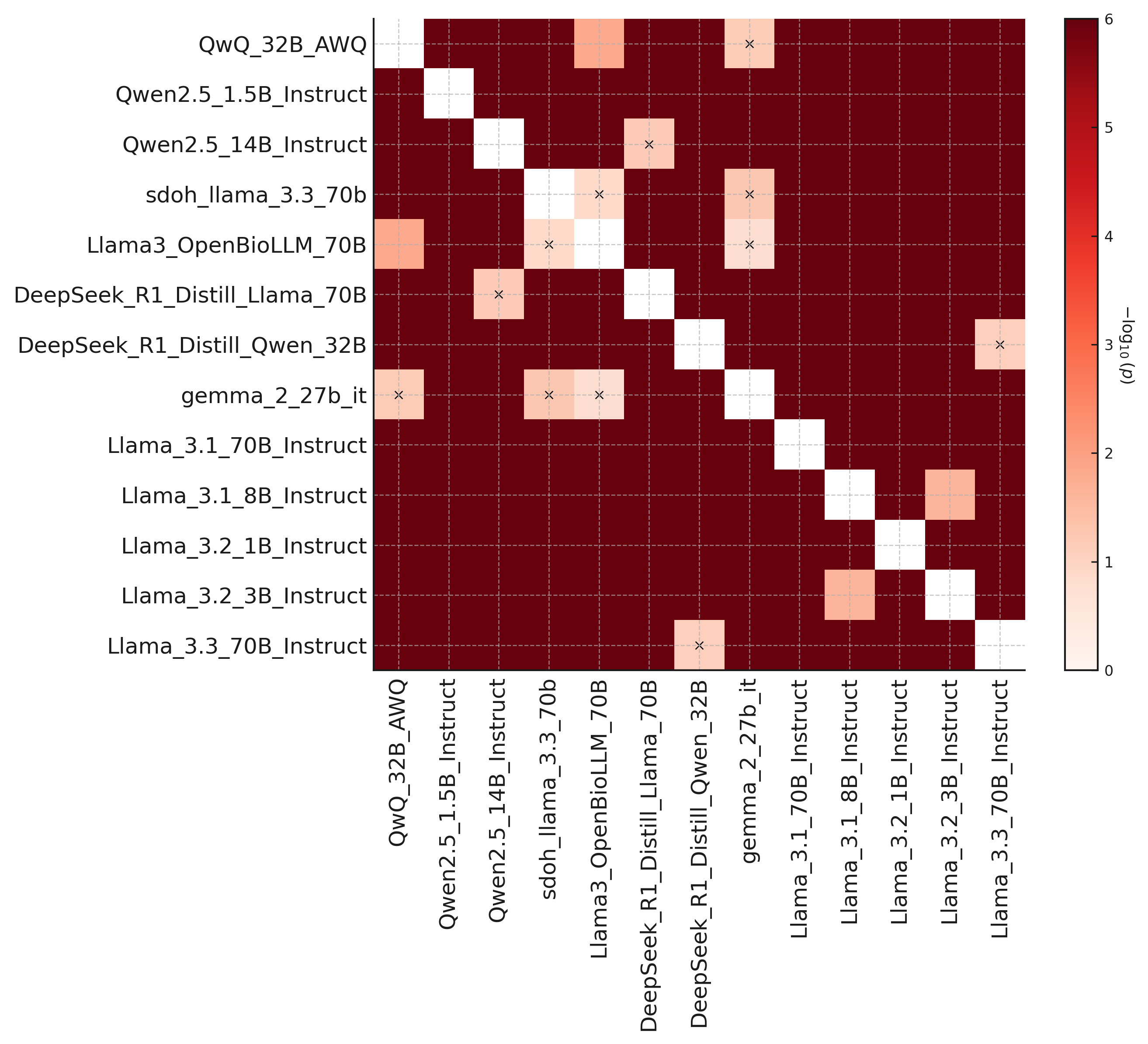}
    \caption*{(c) bootstrap test based on Span F1}
\end{minipage}

\caption{Bootstrap Pairwise significance map between models. Background color encodes $-\log_{10}(p)$ from bootstrap tests (darker cells indicate smaller p-values). Cells marked with ``x'' correspond to pairs that are not deemed statistically significant according to our decision rule.}
\label{fig:bootstrap_maps}
\end{figure*}

As shown in Figure~\ref{fig:bootstrap_maps}, across the three evaluation settings, the significance maps reveal clear differences in model separability. On the code task, high-capacity models (e.g., Llama-3.3-70B, DeepSeek-R1-Distill) exhibit consistently strong statistical advantages, with most pairwise comparisons showing highly significant gaps. In contrast, the span task shows several regions of non-significance among mid-size models, indicating that many systems perform similarly despite small numerical differences. The subcode task is the least separable: a large number of model pairs are statistically tied, suggesting that fine-grained subcode distinctions remain challenging for current LLMs. Overall, larger models tend to dominate, but meaningful performance gaps diminish substantially as tasks become more fine-grained.

Finally, domain-specific models do not always outperform general ones. For instance, {sdoh-llama-3-3-70b}, trained on socially grounded narratives, outperforms biomedical models, suggesting that exposure to discourse-relevant data matters more than domain specialization. In summary, \textsc{EPPCMinerBen} offers a comprehensive framework for evaluating LLM in key dimensions: generalization, alignment, reasoning, and contextual grounding, and serves as a rigorous benchmark to advance clinical communication modeling.

\
\subsection{Ablation Analysis}
To evaluate the design of \textsc{EPPCMinerBen}, we conduct two zero-shot ablation experiments using {Llama-3.1-70B-Instruct} and {Llama-3.3-70B-Instruct} (Table~\ref{fig:ablation}). The first merges code and subcode prediction into a joint task to assess the benefit of label decoupling. The second removes reasoning instructions from prompts to test the role of task-specific guidance.





Ablation results reveal two key insights about task design in \textsc{\textsc{EPPCMinerBen}}. First, predicting code and subcode jointly rather than separately leads to consistent performance drops across all tasks, with notable declines in F1 for both code and subcode classification. This confirms the advantage of hierarchical modeling, which helps models first capture coarse communicative intent before resolving finer distinctions, avoiding semantic confusion introduced by a flattened label space. Second, removing reasoning instructions from prompts significantly harms performance, particularly in evidence extraction, where F1 drops by over 35 points. These findings demonstrate that explicit task guidance is critical for aligning model predictions with discourse-level structures. Together, these results validate our task formulation and emphasize the importance of structural decomposition and prompt design for robust clinical NLP.

\begin{figure}
    \centering
    \includegraphics[width=1\linewidth]{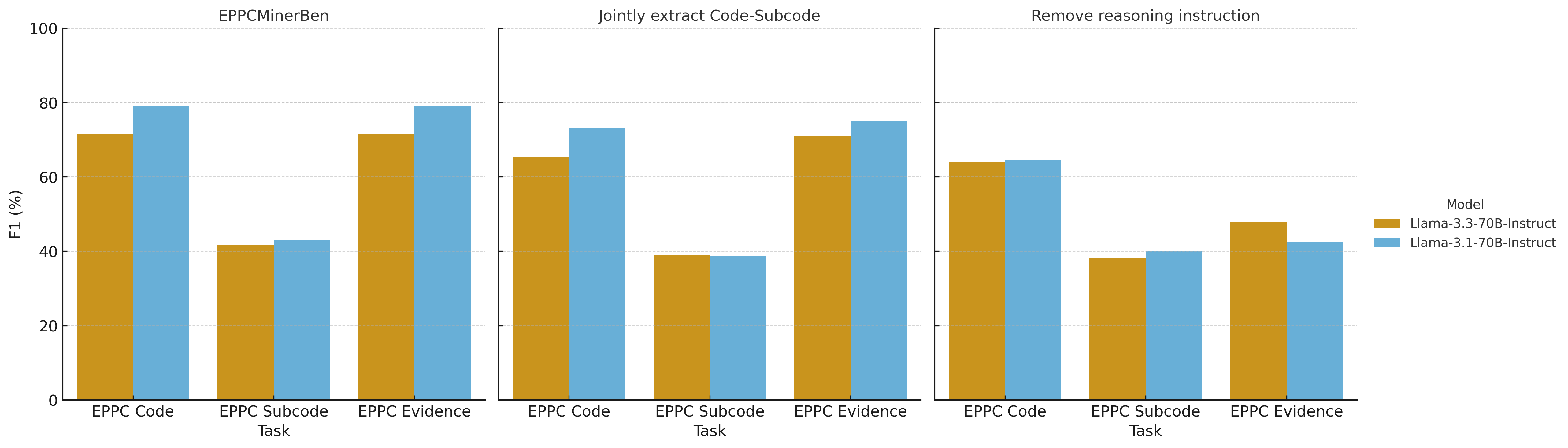}
    \caption{The F1 performance (\%) of ablation study.}
    \label{fig:ablation}
\end{figure}

EPPCMinerBen creates a foundation for multiple future research directions. In particular, it can support advancements in discourse-level modeling of patient–provider interactions, enable investigations into how well models generalize across diverse clinical contexts, and facilitate more rigorous evaluation of patient-centered communication. Together, these contributions help clarify the broader research trajectory enabled by the benchmark and its potential to catalyze subsequent methodological and clinical studies.

\subsection{Limitations}

This study has several limitations that should be acknowledged. First, the dataset is derived from a single healthcare system (Yale New Haven Hospital). Communication styles, linguistic patterns, and patient–provider interaction norms may vary across regions, institutions, and populations, which may limit the generalizability of the benchmark. Second, despite the depth of manual annotation, the dataset’s modest scale (752 messages and 1,933 sentences) may not fully capture the breadth of patient–provider communication, particularly less frequent socio-emotional or culturally specific expressions. Third, the corpus is entirely English-language, which constrains its applicability for studying multilingual or non–English-speaking patient populations who may express needs and concerns differently. Fourth, the benchmark relies on fixed manual prompts and is evaluated only under zero- and few-shot settings, without model fine-tuning. Finally, the study does not include real-world clinical validation, such as assessing how model performance or communication patterns relate to outcomes like treatment adherence, patient satisfaction, or quality of care. Future work should expand the dataset, incorporate multilingual communication, and evaluate the benchmark’s clinical relevance in diverse practice settings.

\section{Conclusion}

In this work, we introduce \textsc{EPPCMinerBen}, a clinically grounded benchmark for evaluating hierarchical classification and evidence extraction in secure patient-provider communication. Our results highlight that instruction-tuned, large-scale models perform well on coarse-grained tasks but still struggle with subcode-level reasoning and token-level evidence alignment. Ablation studies confirm the importance of hierarchical label modeling and reasoning-oriented prompt design. However, the benchmark has limitations, including its reliance on a single institutional dataset, fixed manual prompts, and evaluation under zero-/few-shot settings without fine-tuning. Future work should expand \textsc{EPPCMinerBen} to include multi-institutional and multilingual data, explore adaptive prompting and fine-tuning strategies, and incorporate more nuanced evaluation frameworks such as clinician-in-the-loop assessments. Together, \textsc{EPPCMinerBen} provides a structured, insightful testbed for advancing discourse-aware and reasoning-capable NLP in healthcare.

\section*{Competing Interests}
No competing interest is declared.

\section*{Author Contributions Statement}

S.F. conceptualized the study, designed the methodology, performed the data analysis, and led the writing and revision of the manuscript. Y.W. conducted the experiments, contributed to the analysis, and co-wrote the manuscript. L.M. supported the experimental setup and contributed to data annotation. S.T. performed data annotation and quality assurance. J.A. and S.S. provided domain expertise in patient-provider communication and contributed to the development of the annotation framework. All authors reviewed and approved the final manuscript.

\section*{Acknowledgments}
 This work was supported by the National Cancer Institute [grant number 1R01CA285737-01A1 to S.F.]

\bibliographystyle{elsarticle-num.bst}

\bibliography{elsarticle-ref}

\clearpage
\appendix
\section{Prompt Template}

\begin{center}
  \includegraphics[width=1\linewidth,height=1\textheight,keepaspectratio]{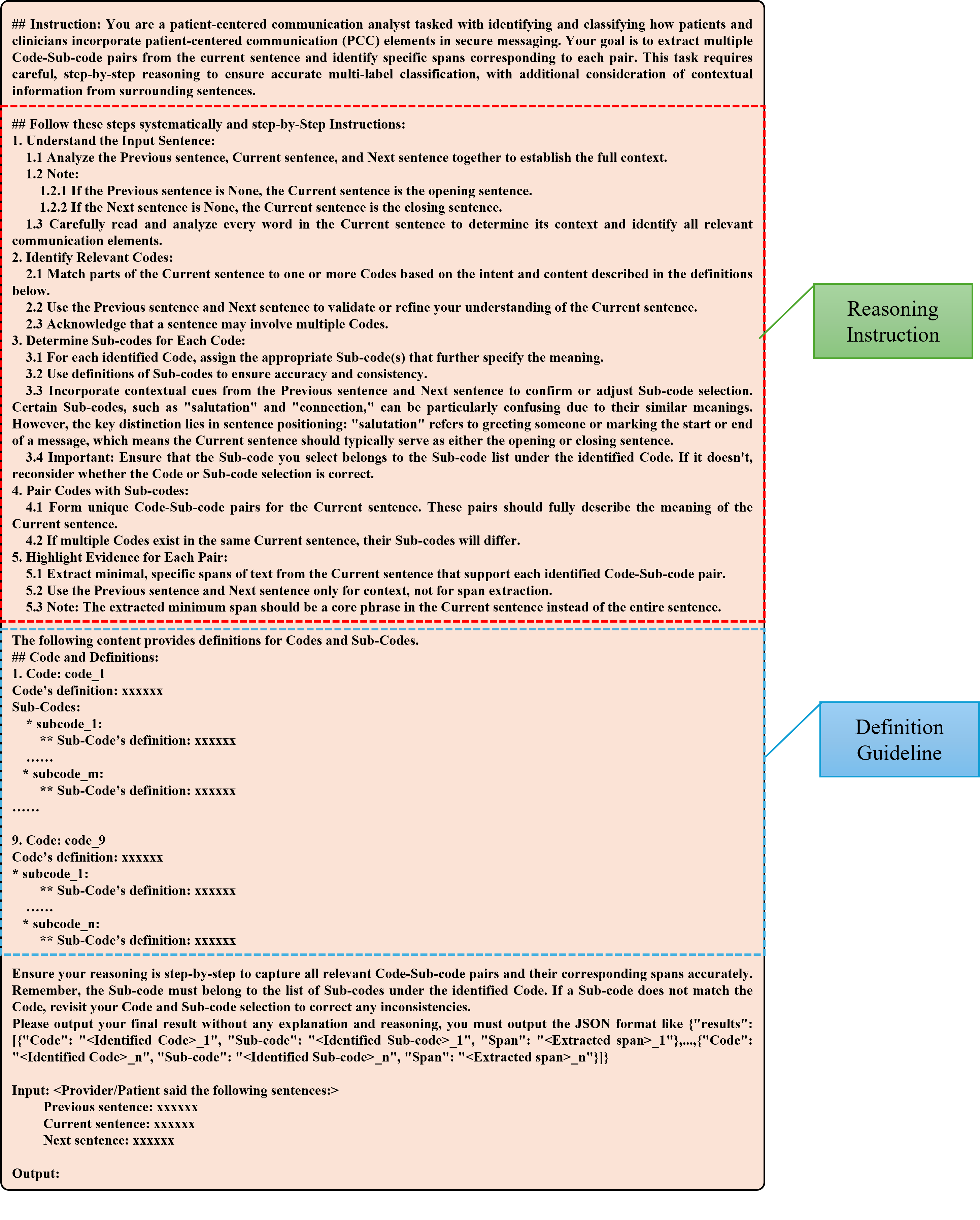}
  \captionof{figure}{The prompt template for EPPCMinerBen.}
  \label{fig:prompt-template}
\end{center}

\newpage
\section{Evaluation Metrics}

We evaluate model performance across the three \textsc{EPPCMinerBen} subtasks using \textbf{micro-averaged Precision}, \textbf{Recall}, and \textbf{F1-score}. Each task adopts a different evaluation strategy tailored to its prediction format and semantic structure:

\paragraph{(1) EPPC Code Classification}

The code classification task is formulated as a multi-label classification problem over a predefined set of communicative codes. Let $\hat{y}_i^{code}$ denote the predicted code set and $y_i^{code}$ is the gold standard. We compute precision recall, and F1-score as follows:

\begin{equation}
    precision_{\text {code }}=\frac{\sum_{i}\left|\hat{y}_{i}^{\text {code }} \cap y_{i}^{\text {code }}\right|}{\sum_{i}\left|\hat{y}_{i}^{\text {code }}\right|}
\end{equation}

\begin{equation}
recall_{\text{code}} = \frac{\sum_i \left| \hat{y}_i^{\text{code}} \cap y_i^{\text{code}} \right|}{\sum_i \left| y_i^{\text{code}} \right|}
\end{equation}

\begin{equation}
F1_{\text{code}} = \frac{2 \times precision_{\text{code}} \times recall_{\text{code}}}{precision_{\text{code}} + recall_{\text{code}}}
\end{equation}

\paragraph{(2) EPPC Subcode Classification}

Subcode classification is also evaluated as a multi-label task, where each message may be annotated with one or more subcodes tied to a parent code. Let $\hat{y}_i^{sub}$ and $y_i^{sub}$ denote predicted and gold subcodes, respectively. Metrics are calculated using:

\begin{equation}
    precision_{\text{sub}} = \frac{\sum_{i} \left| \hat{y}_{i}^{\text{sub}} \cap y_{i}^{\text{sub}} \right|}{\sum_{i} \left| \hat{y}_{i}^{\text{sub}} \right|}
\end{equation}

\begin{equation}
    recall_{\text{sub}} = \frac{\sum_{i} \left| \hat{y}_{i}^{\text{sub}} \cap y_{i}^{\text{sub}} \right|}{\sum_{i} \left| y_{i}^{\text{sub}} \right|}
\end{equation}

\begin{equation}
    F1_{\text{sub}} = \frac{2 \times precision_{\text{sub}} \times recall_{\text{sub}}}{precision_{\text{sub}} + recall_{\text{sub}}}
\end{equation}

\paragraph{(3) EPPC Evidence Extraction}

For span extraction, we evaluate each predicted evidence string against gold spans using a \textbf{relaxed token-level matching strategy}, which combines:
\begin{itemize}
    \item Full containment (i.e., gold span is fully included in predicted span or vice versa).
    \item Jaccard similarity (between predicted and gold spans), with a threshold of 0.6.
\end{itemize}

Let $\hat{e}_i$ and $e_i$ denote the sets of predicted and gold evidence spans for instance $i$. 
A predicted span $\hat{e} \in \hat{e}_i$ is counted as a \textbf{true positive (TP)} if it matches any gold span $e \in e_i$ such that: 
$\textit{Tokens}(e) \subseteq \textit{Tokens}(\hat{e})$ or 
$\textit{Tokens}(\hat{e}) \subseteq \textit{Tokens}(e)$ or 
$\textit{Jaccard}(\hat{e}, e) \geq 0.6$. 
Spans in $\hat{e}_i$ that fail to match any gold span are counted as \textbf{false positives (FP)}, 
and spans in $e_i$ not matched by any prediction are counted as \textbf{false negatives (FN)}. 
Precision, recall, and F1-score are then computed as:

\begin{equation}
    precision_{\text{span}} = \frac{|\text{TP}|}{|\text{TP} + \text{FP}|}
\end{equation}

\begin{equation}
    recall_{\text{span}} = \frac{|\text{TP}|}{|\text{TP} + \text{FN}|}
\end{equation}

\begin{equation}
    F1_{\text{span}} = \frac{2 \times precision_{\text{span}} \times recall_{\text{span}}}{precision_{\text{span}} + recall_{\text{span}}}
\end{equation}

\section{Details of Experiment Results}
\subsection{EPPC Code Classification}

Table~\ref{tab:eppc_code_performance1} details the Precision, Recall, and F1 scores for EPPC code classification. Breaking down the performance into precision and recall reveals distinct behavioral patterns across model architectures that F1 scores alone mask.

\paragraph{Precision-Recall Trade-offs in Large Models}
Top-performing models exhibit different strategies. {LLaMA-3.3-70B-Instruct} achieves its best performance through an "aggressive" strategy in the 0-shot setting, characterized by high Recall (72.43\%) but moderate Precision (57.67\%). Introducing few-shot examples acts as a regularizer: it significantly boosts Precision to 65.75\% (in 2-shot) while slightly tempering Recall, resulting in the highest overall F1 of \textbf{67.03\%}. Conversely, {QwQ-32} adopts a highly "conservative" strategy. It maintains exceptional Precision ($\sim$75\%) across settings, the highest among all models, but suffers from low Recall ($\sim$54\%), indicating it only predicts codes when extremely confident, likely a side effect of its reasoning-heavy optimization.

\paragraph{Impact of Distillation and Scale}
{DeepSeek-R1-Distill-Qwen-32B} shows the most balanced profile, with both Precision and Recall growing steadily with more shots, culminating in a 66.45\% F1 that rivals the 70B models. This suggests the distillation process successfully transferred both the knowledge to identify codes (Recall) and the discrimination to reject false positives (Precision).

\paragraph{Instability in Smaller Models}
The breakdown exposes severe issues in lightweight models. {Llama-3.2-3B-Instruct} in the 0-shot setting achieves a high Recall of 64.36\% (comparable to 70B models) but poor Precision of 19.81\%, suggesting it indiscriminately assigns labels. Few-shot prompting helps curb this behavior, raising Precision to 37.25\%, but it remains insufficient for practical use. {Llama-3.1-8B-Instruct} confirms its formatting failure in 0-shot (0.00 across all metrics) but stabilizes to a balanced $\sim$50\% Precision/Recall in few-shot settings.

\paragraph{Negative Transfer in Gemma:}
For {Gemma-2-27b-it}, the data clarifies why performance drops with examples. In the 0-shot setting, it balances Precision (62.22\%) and Recall (67.21\%) well. However, in few-shot settings, \textit{both} metrics degrade significantly (Precision drops to $\sim$52\%, Recall to $\sim$53\%), confirming that in-context examples systematically confuse the model's decision boundary rather than just biasing it towards one metric.

\begin{table*}[t]
\scriptsize
\centering
\setlength{\tabcolsep}{3pt}
\caption{Performance (\%) for EPPC code classification under different shot settings.\label{tab:eppc_code_performance1}}
\resizebox{\columnwidth}{!}{
\begin{tabular}{l *{9}{c}}
\toprule
\textbf{Model} &
\multicolumn{3}{c}{\textbf{0 shot}} &
\multicolumn{3}{c}{\textbf{1 shot}} &
\multicolumn{3}{c}{\textbf{2 shots}} \\
\cmidrule(lr){2-4} \cmidrule(lr){5-7} \cmidrule(lr){8-10}
& \textbf{P} & \textbf{R} & \textbf{F1}
& \textbf{P} & \textbf{R} & \textbf{F1}
& \textbf{P} & \textbf{R} & \textbf{F1} \\
\midrule

LLaMA-3.3-70B-Instruct 
& \makecell{57.67 ± \\ 0.00e+00}
& \makecell{72.43 ± \\ 0.00e+00}
& \makecell{64.21 ± \\ 0.00e+00}
& \makecell{60.66 ± \\ 2.88e-01}
& \makecell{69.08 ± \\ 1.60e-02}
& \makecell{64.60 ± \\ 1.56e-01}
& \makecell{65.75 ± \\ 6.40e-02}
& \makecell{68.37 ± \\ 1.32e-01}
& \makecell{67.03 ± \\ 1.00e-01} \\
\hline
LLaMA-3.1-70B-Instruct
& \makecell{64.53 ± \\ 4.80e-02}
& \makecell{64.61 ± \\ 1.32e-01}
& \makecell{64.57 ± \\ 9.20e-02}
& \makecell{65.76 ± \\ 7.60e-02}
& \makecell{64.12 ± \\ 8.40e-02}
& \makecell{64.93 ± \\ 8.00e-02}
& \makecell{67.02 ± \\ 2.40e-02}
& \makecell{63.87 ± \\ 0.00e+00}
& \makecell{65.40 ± \\ 1.20e-02} \\
\hline
DeepSeek-R1-Distill-Llama-70B
& \makecell{59.84 ± \\ 3.96e-01}
& \makecell{59.04 ± \\ 1.33e-01}
& \makecell{59.43 ± \\ 2.43e-01}
& \makecell{60.57 ± \\ 0.00e+00}
& \makecell{56.73 ± \\ 0.00e+00}
& \makecell{58.58 ± \\ 0.00e+00}
& \makecell{62.59 ± \\ 0.00e+00}
& \makecell{61.21 ± \\ 1.36e-01}
& \makecell{61.89 ± \\ 6.40e-02} \\
\hline
Gemma-2-27b-it
& \makecell{62.22 ± \\ 0.00e+00}
& \makecell{67.21 ± \\ 0.00e+00}
& \makecell{64.62 ± \\ 0.00e+00}
& \makecell{52.66 ± \\ 0.00e+00}
& \makecell{53.80 ± \\ 0.00e+00}
& \makecell{53.22 ± \\ 0.00e+00}
& \makecell{51.86 ± \\ 0.00e+00}
& \makecell{53.51 ± \\ 0.00e+00}
& \makecell{52.67 ± \\ 0.00e+00} \\
\hline
DeepSeek-R1-Distill-Qwen-32B
& \makecell{60.47 ± \\ 0.00e+00}
& \makecell{62.61 ± \\ 0.00e+00}
& \makecell{61.52 ± \\ 0.00e+00}
& \makecell{63.34 ± \\ 0.00e+00}
& \makecell{65.96 ± \\ 0.00e+00}
& \makecell{64.62 ± \\ 0.00e+00}
& \makecell{65.50 ± \\ 0.00e+00}
& \makecell{67.42 ± \\ 0.00e+00}
& \makecell{66.45 ± \\ 0.00e+00} \\
\hline
QwQ-32
& \makecell{73.16 ± \\ 0.00e+00}
& \makecell{51.13 ± \\ 0.00e+00}
& \makecell{60.19 ± \\ 0.00e+00}
& \makecell{75.09 ± \\ 0.00e+00}
& \makecell{52.76 ± \\ 0.00e+00}
& \makecell{61.97 ± \\ 0.00e+00}
& \makecell{75.17 ± \\ 0.00e+00}
& \makecell{54.76 ± \\ 0.00e+00}
& \makecell{63.36 ± \\ 0.00e+00} \\
\hline
Qwen2.5-14B-Instruct-1M
& \makecell{62.95 ± \\ 2.80e-01}
& \makecell{58.48 ± \\ 1.50e-01}
& \makecell{60.63 ± \\ 2.10e-01}
& \makecell{62.79 ± \\ 1.87e-01}
& \makecell{59.15 ± \\ 1.83e-01}
& \makecell{60.92 ± \\ 1.66e-01}
& \makecell{62.68 ± \\ 1.44e-01}
& \makecell{58.62 ± \\ 1.14e+00}
& \makecell{60.58 ± \\ 6.79e-01} \\
\hline
Llama-3.1-8B-Instruct
& \makecell{0.00 ± \\ 0.00e+00}
& \makecell{0.00 ± \\ 0.00e+00}
& \makecell{0.00 ± \\ 0.00e+00}
& \makecell{49.11 ± \\ 0.00e+00}
& \makecell{53.22 ± \\ 0.00e+00}
& \makecell{51.08 ± \\ 0.00e+00}
& \makecell{49.11 ± \\ 0.00e+00}
& \makecell{53.22 ± \\ 0.00e+00}
& \makecell{51.08 ± \\ 0.00e+00} \\
\hline
Llama-3.2-3B-Instruct
& \makecell{19.81 ± \\ 2.43e-01}
& \makecell{64.36 ± \\ 5.67e-01}
& \makecell{30.30 ± \\ 2.42e-01}
& \makecell{37.25 ± \\ 0.00e+00}
& \makecell{42.65 ± \\ 0.00e+00}
& \makecell{39.77 ± \\ 0.00e+00}
& \makecell{37.25 ± \\ 0.00e+00}
& \makecell{42.65 ± \\ 0.00e+00}
& \makecell{39.77 ± \\ 0.00e+00} \\
\hline
Llama-3.2-1B-Instruct
& \makecell{22.66 ± \\ 1.17e-01}
& \makecell{33.89 ± \\ 9.60e-02}
& \makecell{27.16 ± \\ 9.70e-02}
& \makecell{18.58 ± \\ 0.00e+00}
& \makecell{38.01 ± \\ 0.00e+00}
& \makecell{24.96 ± \\ 0.00e+00}
& \makecell{18.58 ± \\ 0.00e+00}
& \makecell{38.01 ± \\ 0.00e+00}
& \makecell{24.96 ± \\ 0.00e+00} \\
\hline
Qwen2.5-1.5B-Instruct
& \makecell{25.11 ± \\ 2.20e-02}
& \makecell{26.69 ± \\ 7.10e-02}
& \makecell{25.88 ± \\ 3.80e-02}
& \makecell{42.40 ± \\ 5.55e-15}
& \makecell{34.50 ± \\ 0.00e+00}
& \makecell{38.05 ± \\ 0.00e+00}
& \makecell{42.40 ± \\ 5.55e-15}
& \makecell{34.50 ± \\ 0.00e+00}
& \makecell{38.05 ± \\ 0.00e+00} \\
\hline
Llama3-OpenBioLLM-70B
& \makecell{55.87 ± \\ 0.00e+00}
& \makecell{47.70 ± \\ 0.00e+00}
& \makecell{51.46 ± \\ 0.00e+00}
& \makecell{55.68 ± \\ 0.00e+00}
& \makecell{50.13 ± \\ 0.00e+00}
& \makecell{52.76 ± \\ 0.00e+00}
& \makecell{55.58 ± \\ 0.00e+00}
& \makecell{49.96 ± \\ 5.55e-15}
& \makecell{52.62 ± \\ 0.00e+00} \\
\hline
sdoh-llama-3-3-70b
& \makecell{64.65 ± \\ 0.00e+00}
& \makecell{62.57 ± \\ 0.00e+00}
& \makecell{63.60 ± \\ 0.00e+00}
& \makecell{66.40 ± \\ 0.00e+00}
& \makecell{61.49 ± \\ 0.00e+00}
& \makecell{63.85 ± \\ 0.00e+00}
& \makecell{66.42 ± \\ 0.00e+00}
& \makecell{63.03 ± \\ 0.00e+00}
& \makecell{64.68 ± \\ 0.00e+00} \\

\bottomrule
\end{tabular}
}
\end{table*}

\subsection{EPPC Subcode Classification}

{Table~\ref{tab:eppc_subcode_performance1} provides a granular breakdown of Precision, Recall, and F1 scores for the subcode classification task. This finer-grained classification challenge exacerbates the strategic differences between models, revealing critical insights into their decision boundaries.

\paragraph{The Precision-Recall Trade-off in Reasoning Models}
{QwQ-32} demonstrates a distinct "high-precision" personality. Across all settings, it maintains the highest Precision (peaking at \textbf{56.64\%} in 2-shot) but struggles with Recall ($\sim$40\%). This indicates that the model's reasoning process makes it highly conservative; it only predicts a specific subcode when the evidence is overwhelming, resulting in many missed detections (false negatives) but few false positives.

\paragraph{Calibration in General-Purpose Models}
In contrast, {LLaMA-3.3-70B-Instruct} adopts a "high-recall" strategy in the 0-shot setting (Recall 52.29\%, Precision 34.68\%), effectively casting a wide net to capture potential intents. However, this comes at the cost of precision, suggesting the model struggles to discriminate between semantically similar subcodes without examples. Few-shot prompting significantly aids in calibrating this behavior, raising Precision to 42.53\% in the 2-shot setting while normalizing Recall.

\paragraph{Superior Balance via Distillation}
{DeepSeek-R1-Distill-Qwen-32B} achieves the best overall F1 (48.25\%) by striking a superior balance between Precision and Recall (both hovering around 47--49\%). Unlike Llama-3.3 (which trades Precision for Recall) or QwQ (which trades Recall for Precision), the distilled DeepSeek model exhibits robust discrimination capabilities, correctly identifying subcodes without being overly conservative or reckless.

\paragraph{Breakdown of Lightweight Models}
The Precision/Recall split exposes the severity of the failure in smaller models. {Llama-3.2-3B-Instruct} in 0-shot achieves a Recall of 33.38\% but a Precision of only \textbf{7.54\%}. This extreme imbalance indicates "hallucinatory labelling", the model appears to assign subcodes without discrimination, yielding a few correct predictions largely by chance while generating many false positives.
Similarly, {Gemma-2-27b-it} shows a simultaneous collapse in both Precision (38.65\% $\to$ 30.22\%) and Recall (46.29\% $\to$ 31.56\%) when moving from 0-shot to 2-shot, confirming that few-shot examples actively degrade its internal representation of the subcode semantics.

\begin{table*}[t]
\scriptsize
\centering
\setlength{\tabcolsep}{3pt}
\caption{Performance (\%) for EPPC subcode classification under different shot settings.\label{tab:eppc_subcode_performance1}}
\resizebox{\columnwidth}{!}{
\begin{tabular}{l *{9}{c}}
\toprule
\textbf{Model} &
\multicolumn{3}{c}{\textbf{0 shot}} &
\multicolumn{3}{c}{\textbf{1 shot}} &
\multicolumn{3}{c}{\textbf{2 shots}} \\
\cmidrule(lr){2-4} \cmidrule(lr){5-7} \cmidrule(lr){8-10}
& \textbf{P} & \textbf{R} & \textbf{F1}
& \textbf{P} & \textbf{R} & \textbf{F1}
& \textbf{P} & \textbf{R} & \textbf{F1} \\
\midrule

LLaMA-3.3-70B-Instruct
& \makecell{34.68 $\pm$ \\ 0.00e+00}
& \makecell{52.29 $\pm$ \\ 0.00e+00}
& \makecell{41.70 $\pm$ \\ 0.00e+00}
& \makecell{37.60 $\pm$ \\ 1.76e-01}
& \makecell{48.06 $\pm$ \\ 1.60e-02}
& \makecell{42.19 $\pm$ \\ 1.16e-01}
& \makecell{42.53 $\pm$ \\ 1.48e-01}
& \makecell{46.58 $\pm$ \\ 1.60e-02}
& \makecell{44.46 $\pm$ \\ 8.40e-02} \\
\hline
LLaMA-3.1-70B-Instruct
& \makecell{40.19 $\pm$ \\ 1.16e-01}
& \makecell{46.40 $\pm$ \\ 0.00e+00}
& \makecell{43.08 $\pm$ \\ 6.80e-02}
& \makecell{44.15 $\pm$ \\ 4.00e-02}
& \makecell{44.29 $\pm$ \\ 4.40e-02}
& \makecell{44.22 $\pm$ \\ 4.00e-02}
& \makecell{45.23 $\pm$ \\ 1.20e-02}
& \makecell{43.12 $\pm$ \\ 3.20e-02}
& \makecell{44.15 $\pm$ \\ 2.00e-02} \\
\hline
DeepSeek-R1-Distill-Llama-70B
& \makecell{48.04 $\pm$ \\ 2.65e-01}
& \makecell{48.30 $\pm$ \\ 2.36e-01}
& \makecell{48.17 $\pm$ \\ 1.52e-01}
& \makecell{45.83 $\pm$ \\ 5.55e-15}
& \makecell{44.04 $\pm$ \\ 0.00e+00}
& \makecell{44.92 $\pm$ \\ 0.00e+00}
& \makecell{46.19 $\pm$ \\ 6.80e-02}
& \makecell{46.11 $\pm$ \\ 1.36e-01}
& \makecell{46.15 $\pm$ \\ 3.60e-02} \\
\hline
Gemma-2-27b-it
& \makecell{38.65 $\pm$ \\ 0.00e+00}
& \makecell{46.29 $\pm$ \\ 0.00e+00}
& \makecell{42.13 $\pm$ \\ 0.00e+00}
& \makecell{31.04 $\pm$ \\ 0.00e+00}
& \makecell{32.65 $\pm$ \\ 0.00e+00}
& \makecell{31.82 $\pm$ \\ 0.00e+00}
& \makecell{30.22 $\pm$ \\ 0.00e+00}
& \makecell{31.56 $\pm$ \\ 0.00e+00}
& \makecell{30.88 $\pm$ \\ 0.00e+00} \\
\hline
DeepSeek-R1-Distill-Qwen-32B
& \makecell{47.22 $\pm$ \\ 5.55e-15}
& \makecell{49.33 $\pm$ \\ 5.55e-15}
& \makecell{48.25 $\pm$ \\ 5.55e-15}
& \makecell{46.27 $\pm$ \\ 0.00e+00}
& \makecell{49.59 $\pm$ \\ 5.55e-15}
& \makecell{47.87 $\pm$ \\ 0.00e+00}
& \makecell{46.92 $\pm$ \\ 0.00e+00}
& \makecell{49.03 $\pm$ \\ 5.55e-15}
& \makecell{47.95 $\pm$ \\ 0.00e+00} \\
\hline
QwQ-32
& \makecell{55.35 $\pm$ \\ 0.00e+00}
& \makecell{38.42 $\pm$ \\ 0.00e+00}
& \makecell{45.35 $\pm$ \\ 0.00e+00}
& \makecell{56.24 $\pm$ \\ 0.00e+00}
& \makecell{38.83 $\pm$ \\ 0.00e+00}
& \makecell{45.94 $\pm$ \\ 5.55e-15}
& \makecell{56.64 $\pm$ \\ 0.00e+00}
& \makecell{40.14 $\pm$ \\ 5.55e-15}
& \makecell{46.98 $\pm$ \\ 5.55e-15} \\
\hline
Qwen2.5-14B-Instruct-1M
& \makecell{42.96 $\pm$ \\ 6.74e-02}
& \makecell{42.14 $\pm$ \\ 8.96e-02}
& \makecell{42.55 $\pm$ \\ 7.05e-02}
& \makecell{41.92 $\pm$ \\ 1.73e-01}
& \makecell{41.73 $\pm$ \\ 1.51e-01}
& \makecell{41.83 $\pm$ \\ 1.53e-01}
& \makecell{41.90 $\pm$ \\ 3.34e-01}
& \makecell{41.24 $\pm$ \\ 1.38e+00}
& \makecell{41.56 $\pm$ \\ 8.74e-01} \\
\hline
Llama-3.1-8B-Instruct
& \makecell{0.00 $\pm$ \\ 0.00e+00}
& \makecell{0.00 $\pm$ \\ 0.00e+00}
& \makecell{0.00 $\pm$ \\ 0.00e+00}
& \makecell{19.59 $\pm$ \\ 0.00e+00}
& \makecell{33.25 $\pm$ \\ 0.00e+00}
& \makecell{24.66 $\pm$ \\ 0.00e+00}
& \makecell{19.59 $\pm$ \\ 0.00e+00}
& \makecell{33.25 $\pm$ \\ 0.00e+00}
& \makecell{24.66 $\pm$ \\ 0.00e+00} \\
\hline
Llama-3.2-3B-Instruct
& \makecell{7.54 $\pm$ \\ 2.58e-02}
& \makecell{33.38 $\pm$ \\ 4.41e-01}
& \makecell{12.31 $\pm$ \\ 4.58e-02}
& \makecell{12.19 $\pm$ \\ 1.39e-15}
& \makecell{21.85 $\pm$ \\ 0.00e+00}
& \makecell{15.65 $\pm$ \\ 0.00e+00}
& \makecell{12.19 $\pm$ \\ 1.39e-15}
& \makecell{21.85 $\pm$ \\ 0.00e+00}
& \makecell{15.65 $\pm$ \\ 0.00e+00} \\
\hline
Llama-3.2-1B-Instruct
& \makecell{3.54 $\pm$ \\ 3.88e-02}
& \makecell{10.22 $\pm$ \\ 1.86e-01}
& \makecell{5.26 $\pm$ \\ 6.97e-02}
& \makecell{5.48 $\pm$ \\ 0.00e+00}
& \makecell{16.79 $\pm$ \\ 0.00e+00}
& \makecell{8.26 $\pm$ \\ 0.00e+00}
& \makecell{5.48 $\pm$ \\ 0.00e+00}
& \makecell{16.79 $\pm$ \\ 0.00e+00}
& \makecell{8.26 $\pm$ \\ 0.00e+00} \\
\hline
Qwen2.5-1.5B-Instruct
& \makecell{7.10 $\pm$ \\ 1.22e-01}
& \makecell{7.39 $\pm$ \\ 1.02e-01}
& \makecell{7.24 $\pm$ \\ 1.09e-01}
& \makecell{15.27 $\pm$ \\ 0.00e+00}
& \makecell{13.76 $\pm$ \\ 0.00e+00}
& \makecell{14.47 $\pm$ \\ 0.00e+00}
& \makecell{15.27 $\pm$ \\ 0.00e+00}
& \makecell{13.76 $\pm$ \\ 0.00e+00}
& \makecell{14.47 $\pm$ \\ 0.00e+00} \\
\hline
Llama3-OpenBioLLM-70B
& \makecell{30.96 $\pm$ \\ 0.00e+00}
& \makecell{27.32 $\pm$ \\ 0.00e+00}
& \makecell{29.03 $\pm$ \\ 0.00e+00}
& \makecell{28.67 $\pm$ \\ 0.00e+00}
& \makecell{25.71 $\pm$ \\ 0.00e+00}
& \makecell{27.11 $\pm$ \\ 0.00e+00}
& \makecell{28.33 $\pm$ \\ 0.00e+00}
& \makecell{25.56 $\pm$ \\ 0.00e+00}
& \makecell{26.88 $\pm$ \\ 0.00e+00} \\
\hline
sdoh-llama-3-3-70b
& \makecell{40.86 $\pm$ \\ 0.00e+00}
& \makecell{40.74 $\pm$ \\ 0.00e+00}
& \makecell{40.80 $\pm$ \\ 5.55e-15}
& \makecell{42.63 $\pm$ \\ 0.00e+00}
& \makecell{39.99 $\pm$ \\ 0.00e+00}
& \makecell{41.27 $\pm$ \\ 5.55e-15}
& \makecell{43.43 $\pm$ \\ 0.00e+00}
& \makecell{41.60 $\pm$ \\ 5.55e-15}
& \makecell{42.50 $\pm$ \\ 0.00e+00} \\

\bottomrule
\end{tabular}
}
\end{table*}

\subsection{EPPC Evidence Extraction}

Table~\ref{tab:eppc_evidence_performance1} breaks down the performance into Precision and Recall, revealing extraction behaviors that are obscured by aggregated F1 scores.

\paragraph{Evidence Extraction Difficulties in High-Capacity Models}
{LLaMA-3.3-70B-Instruct} reveals a striking behavior in the 0-shot setting: it achieves a near-perfect Recall of \textbf{89.82\%}, the highest in the entire benchmark, but a mediocre Precision of 58.72\%. This indicates a tendency towards "over-extraction," where the model correctly finds the evidence but includes excessive surrounding context or irrelevant text. The introduction of few-shot examples acts as a constraint mechanism, significantly boosting Precision to 74.29\% (2-shot) by teaching the model to predict tighter, more concise spans, albeit with a slight drop in Recall.

\paragraph{Stability in Extraction}
{LLaMA-3.1-70B-Instruct} proves to be the most well-calibrated model for this task. Unlike its 3.3 counterpart, it maintains high Precision ($\sim$82--85\%) and high Recall ($\sim$80--84\%) across all settings without needing examples to correct its behavior. This suggests Llama-3.1's instruction tuning may have emphasized precise grounding more effectively than Llama-3.3.

\paragraph{Conservative Extraction in Reasoning Models}
{QwQ-32} continues to exhibit very cautious predictions. It exhibits high Precision ($\sim$79\%) but remarkably low Recall ($\sim$55\%). In an extraction context, this means the model often returns incomplete spans or refuses to extract evidence unless the signal is unambiguous, leading to high "false negative" rates  despite good accuracy on the predictions it does make.

\paragraph{Evidence Extraction Difficulties in Smaller Models}

The performance breakdown clarifies why smaller models such as Llama-3.1-8B-Instruct and Llama-3.2-3B struggle. In the zero-shot setting, they show very high Recall (86.08\% for 8B, 70.55\% for 3B) but extremely Precision (27.04\% and 8.55\%, respectively). This imbalance indicates that the models tend to reproduce large portions of the input text in an effort to capture relevant evidence, rather than conducting targeted or semantically informed segmentation.

\paragraph{Ideally Balanced Distillation}
{DeepSeek-R1-Distill-Qwen-32B} demonstrates excellent calibration. With Precision and Recall both consistently tracking in the mid-to-high 70s, it avoids both the over-extraction of Llama-3.3 and the under-extraction of QwQ. This balance enables it to rival 70B models in overall F1, proving that distilled models can retain precise grounding capabilities.

\begin{table*}[t]
\scriptsize
\centering
\setlength{\tabcolsep}{3pt}
\caption{Performance (\%) based on Jaccard similarity for EPPC evidence extraction under different shot settings.\label{tab:eppc_evidence_performance1}}
\resizebox{\columnwidth}{!}{%
\begin{tabular}{l *{9}{c}}
\toprule
\textbf{Model} &
\multicolumn{3}{c}{\textbf{0 shot}} &
\multicolumn{3}{c}{\textbf{1 shot}} &
\multicolumn{3}{c}{\textbf{2 shots}} \\
\cmidrule(lr){2-4} \cmidrule(lr){5-7} \cmidrule(lr){8-10}
& \textbf{P} & \textbf{R} & \textbf{F1}
& \textbf{P} & \textbf{R} & \textbf{F1}
& \textbf{P} & \textbf{R} & \textbf{F1} \\
\midrule

LLaMA-3.3-70B-Instruct
& \makecell{58.72 $\pm$ \\ 0.00e+00}
& \makecell{89.82 $\pm$ \\ 1.11e-14}
& \makecell{71.01 $\pm$ \\ 0.00e+00}
& \makecell{54.98 $\pm$ \\ 6.04e-01}
& \makecell{85.35 $\pm$ \\ 7.20e-02}
& \makecell{66.88 $\pm$ \\ 4.20e-01}
& \makecell{74.29 $\pm$ \\ 1.42e+00}
& \makecell{84.01 $\pm$ \\ 1.48e-01}
& \makecell{78.84 $\pm$ \\ 7.52e-01} \\
\hline
LLaMA-3.1-70B-Instruct
& \makecell{74.13 $\pm$ \\ 1.92e-01}
& \makecell{84.27 $\pm$ \\ 7.20e-02}
& \makecell{78.87 $\pm$ \\ 7.60e-02}
& \makecell{82.62 $\pm$ \\ 4.00e-03}
& \makecell{81.50 $\pm$ \\ 3.20e-02}
& \makecell{82.06 $\pm$ \\ 2.00e-02}
& \makecell{85.65 $\pm$ \\ 5.20e-02}
& \makecell{80.21 $\pm$ \\ 7.20e-02}
& \makecell{82.84 $\pm$ \\ 6.00e-02} \\
\hline
DeepSeek-R1-Distill-Llama-70B
& \makecell{72.64 $\pm$ \\ 3.83e-01}
& \makecell{73.33 $\pm$ \\ 2.54e-01}
& \makecell{72.98 $\pm$ \\ 1.08e-01}
& \makecell{73.60 $\pm$ \\ 0.00e+00}
& \makecell{71.33 $\pm$ \\ 0.00e+00}
& \makecell{72.45 $\pm$ \\ 0.00e+00}
& \makecell{74.16 $\pm$ \\ 2.60e-01}
& \makecell{73.65 $\pm$ \\ 0.00e+00}
& \makecell{73.90 $\pm$ \\ 1.28e-01} \\
\hline
Gemma-2-27b-it
& \makecell{61.97 $\pm$ \\ 0.00e+00}
& \makecell{80.21 $\pm$ \\ 0.00e+00}
& \makecell{69.92 $\pm$ \\ 0.00e+00}
& \makecell{62.63 $\pm$ \\ 0.00e+00}
& \makecell{72.35 $\pm$ \\ 0.00e+00}
& \makecell{67.14 $\pm$ \\ 0.00e+00}
& \makecell{62.76 $\pm$ \\ 0.00e+00}
& \makecell{72.56 $\pm$ \\ 0.00e+00}
& \makecell{67.31 $\pm$ \\ 0.00e+00} \\
\hline
DeepSeek-R1-Distill-Qwen-32B
& \makecell{74.19 $\pm$ \\ 0.00e+00}
& \makecell{78.14 $\pm$ \\ 0.00e+00}
& \makecell{76.12 $\pm$ \\ 0.00e+00}
& \makecell{72.97 $\pm$ \\ 0.00e+00}
& \makecell{77.89 $\pm$ \\ 0.00e+00}
& \makecell{75.35 $\pm$ \\ 0.00e+00}
& \makecell{75.87 $\pm$ \\ 0.00e+00}
& \makecell{78.65 $\pm$ \\ 0.00e+00}
& \makecell{77.24 $\pm$ \\ 0.00e+00} \\
\hline
QwQ-32
& \makecell{79.58 $\pm$ \\ 0.00e+00}
& \makecell{55.09 $\pm$ \\ 0.00e+00}
& \makecell{65.11 $\pm$ \\ 0.00e+00}
& \makecell{78.09 $\pm$ \\ 0.00e+00}
& \makecell{55.42 $\pm$ \\ 0.00e+00}
& \makecell{64.83 $\pm$ \\ 0.00e+00}
& \makecell{77.66 $\pm$ \\ 0.00e+00}
& \makecell{57.59 $\pm$ \\ 0.00e+00}
& \makecell{66.14 $\pm$ \\ 0.00e+00} \\
\hline
Qwen2.5-14B-Instruct-1M
& \makecell{77.14 $\pm$ \\ 1.53e-01}
& \makecell{76.24 $\pm$ \\ 1.80e-01}
& \makecell{76.69 $\pm$ \\ 1.67e-01}
& \makecell{72.37 $\pm$ \\ 1.45e-01}
& \makecell{75.04 $\pm$ \\ 1.94e-01}
& \makecell{73.68 $\pm$ \\ 1.35e-01}
& \makecell{73.32 $\pm$ \\ 1.86e+00}
& \makecell{74.74 $\pm$ \\ 6.43e-01}
& \makecell{74.00 $\pm$ \\ 6.42e-01} \\
\hline
Llama-3.1-8B-Instruct
& \makecell{27.04 $\pm$ \\ 0.00e+00}
& \makecell{86.08 $\pm$ \\ 0.00e+00}
& \makecell{41.15 $\pm$ \\ 5.55e-15}
& \makecell{24.04 $\pm$ \\ 0.00e+00}
& \makecell{75.10 $\pm$ \\ 0.00e+00}
& \makecell{36.42 $\pm$ \\ 0.00e+00}
& \makecell{24.04 $\pm$ \\ 0.00e+00}
& \makecell{75.10 $\pm$ \\ 0.00e+00}
& \makecell{36.42 $\pm$ \\ 0.00e+00} \\
\hline
Llama-3.2-3B-Instruct
& \makecell{8.55 $\pm$ \\ 7.96e-02}
& \makecell{70.55 $\pm$ \\ 3.08e-01}
& \makecell{15.25 $\pm$ \\ 1.34e-01}
& \makecell{22.30 $\pm$ \\ 0.00e+00}
& \makecell{59.70 $\pm$ \\ 0.00e+00}
& \makecell{32.47 $\pm$ \\ 0.00e+00}
& \makecell{22.30 $\pm$ \\ 0.00e+00}
& \makecell{59.70 $\pm$ \\ 0.00e+00}
& \makecell{32.47 $\pm$ \\ 0.00e+00} \\
\hline
Llama-3.2-1B-Instruct
& \makecell{4.77 $\pm$ \\ 7.48e-03}
& \makecell{62.19 $\pm$ \\ 2.86e-01}
& \makecell{8.86 $\pm$ \\ 1.86e-02}
& \makecell{3.70 $\pm$ \\ 0.00e+00}
& \makecell{44.40 $\pm$ \\ 5.55e-15}
& \makecell{6.84 $\pm$ \\ 0.00e+00}
& \makecell{3.70 $\pm$ \\ 0.00e+00}
& \makecell{44.40 $\pm$ \\ 5.55e-15}
& \makecell{6.84 $\pm$ \\ 0.00e+00} \\
\hline
Qwen2.5-1.5B-Instruct
& \makecell{27.04 $\pm$ \\ 1.80e-01}
& \makecell{54.62 $\pm$ \\ 1.76e-01}
& \makecell{36.17 $\pm$ \\ 1.88e-01}
& \makecell{43.98 $\pm$ \\ 5.55e-15}
& \makecell{58.21 $\pm$ \\ 0.00e+00}
& \makecell{50.10 $\pm$ \\ 0.00e+00}
& \makecell{43.98 $\pm$ \\ 5.55e-15}
& \makecell{58.21 $\pm$ \\ 0.00e+00}
& \makecell{50.10 $\pm$ \\ 0.00e+00} \\
\hline
Llama3-OpenBioLLM-70B
& \makecell{75.10 $\pm$ \\ 0.00e+00}
& \makecell{66.69 $\pm$ \\ 0.00e+00}
& \makecell{70.65 $\pm$ \\ 0.00e+00}
& \makecell{73.68 $\pm$ \\ 0.00e+00}
& \makecell{66.47 $\pm$ \\ 0.00e+00}
& \makecell{69.89 $\pm$ \\ 0.00e+00}
& \makecell{72.76 $\pm$ \\ 0.00e+00}
& \makecell{65.93 $\pm$ \\ 0.00e+00}
& \makecell{69.18 $\pm$ \\ 0.00e+00} \\
\hline
sdoh-llama-3-3-70b
& \makecell{73.30 $\pm$ \\ 0.00e+00}
& \makecell{73.54 $\pm$ \\ 0.00e+00}
& \makecell{73.42 $\pm$ \\ 0.00e+00}
& \makecell{70.01 $\pm$ \\ 0.00e+00}
& \makecell{70.32 $\pm$ \\ 0.00e+00}
& \makecell{70.16 $\pm$ \\ 0.00e+00}
& \makecell{69.09 $\pm$ \\ 0.00e+00}
& \makecell{71.04 $\pm$ \\ 0.00e+00}
& \makecell{70.05 $\pm$ \\ 0.00e+00} \\
\bottomrule
\end{tabular}
}
\end{table*}

\subsection{Evaluation Settings}
We utilize the LM Evaluation Harness v0.4.0, with vLLM v0.10.0 and the PyTorch 2.7.1+cu126 framework\footnote{\url{https://github.com/EleutherAI/lm-evaluation-harness}} to create tailored benchmark suites. We used default settings for hyperparameters. A set of random seeds was used for reproducibility: {1248, 1087, 5674, 2583, 4900}. The batch size was dynamically determined based on available computational resources, and computations were performed using float32 precision.
All models are assessed locally on a 4×H100 GPU cluster (each with 80GB of memory). The input length is standardized to 8,192 tokens for zero-shot evaluation, 16,384 tokens for one-shot, and 24,576 tokens for two-shot settings. The generation length is capped at 1,024 tokens to accommodate reasoning-intensive outputs. The total time spent on evaluation amounts to approximately 450 GPU hours.







\end{document}